%% file: neurips_2026.tex
\documentclass{article}
\PassOptionsToPackage{table}{xcolor}

\usepackage[main, final]{neurips_2026}

\usepackage{booktabs}
\usepackage{longtable}
\usepackage{array}
\usepackage[most]{tcolorbox}
\usepackage{xcolor}
\usepackage{fvextra} 
\usepackage{longtable}
\usepackage{booktabs}
\usepackage{array}
\usepackage{xcolor}

\usepackage{booktabs}      
\usepackage{multirow}      
\usepackage[table]{xcolor} 

\usepackage[utf8]{inputenc} 
\usepackage[T1]{fontenc}    
\usepackage{hyperref}       
\usepackage{url}            
\usepackage{booktabs}       
\usepackage{amsfonts}       
\usepackage{nicefrac}       
\usepackage{microtype}      
\usepackage{xcolor}         
\usepackage{algorithm}
\usepackage{algpseudocode}
\usepackage{enumitem}
\usepackage{amsmath}

\usepackage{graphicx}
\usepackage{booktabs}
\usepackage{multirow}
\usepackage{pifont}
\usepackage{caption,subcaption}
\usepackage{adjustbox}
\usepackage{soul, color}
\definecolor{myorange}{RGB}{255,165,0}

\definecolor{mygray}{RGB}{192,192,192}
\usepackage[most]{tcolorbox}
\usepackage{float}
\usepackage{xspace}
\tcbset{
  aibox/.style={
    width=\textwidth,
    top=10pt,
    colback=white,
    colframe=black,
    colbacktitle=black,
    enhanced,
    center,
    attach boxed title to top center={yshift=-0.1in},
    boxed title style={boxrule=0pt,colframe=white,},
  }
}
\newtcolorbox{AIbox}[2][]{aibox,title=#2,#1}
\usepackage{listings}
\definecolor{myblue}{RGB}{100, 150, 200}
\definecolor{mygreen}{RGB}{80, 160, 80}

\definecolor{darkgreen}{rgb}{0.0, 0.5, 0.0}
\definecolor{darkgray}{gray}{0.4}
\definecolor{maroon}{rgb}{0.5, 0.0, 0.0}
\definecolor{navy}{rgb}{0.0, 0.0, 0.5}
\definecolor{teal}{rgb}{0.0, 0.5, 0.5}

\input{math_commands.tex}

\usepackage{hyperref}
\usepackage{url}
\title{EvoOptiGraph: Weakness-Driven Coevolution via Graph-Based Structural Generation for Optimization Modeling}

\author{
Qingcan Kang$^{1*}$, Mingyang LIU$^{2*}$,\textbf{ Xiaojin Fu}$^{1}$, Shixiong Kai$^{1}$, Tao Zhong$^{1}$, 
~\textbf{Mingxuan Yuan}$^{1\dagger}$\\
     $^{1}$Huawei Noah's Ark Lab, Hong Kong \\
     $^{2}$Department of Computer Science, City University of Hong Kong \\     
     \texttt{\{kangqingcan,fuxiaojin,kaishixiong,zhongtao5,Yuan.Mingxuan\}@huawei.com}\\
    \texttt{mingyaliu8-c@my.cityu.edu.hk}
}

\begin{document}

\maketitle

\def\customfootnotetext#1#2{{%
		\let\thefootnote\relax
		\footnotetext[#1]{#2}}}

\customfootnotetext{1}{\textsuperscript{*} Equal contribution.}
\customfootnotetext{2}{{$\dagger$} Corresponding author.}

\input{00_abstract.tex}
\input{01_introduction.tex}

\input{02_related_work.tex}

\input{03_methods.tex}

\input{04_experiments.tex}

\input{05_conclusion.tex}

\bibliographystyle{plainnat}
\bibliography{06_references}

\newpage
\appendix
\input{07_appendix}

\end{document}

%% file: math_commands.tex
\usepackage{amsmath,amsfonts,bm}

\def\eqref#1{equation~\ref{#1}}

\def\1{\bm{1}}

\DeclareMathAlphabet{\mathsfit}{\encodingdefault}{\sfdefault}{m}{sl}
\SetMathAlphabet{\mathsfit}{bold}{\encodingdefault}{\sfdefault}{bx}{n}



%% file: 00_abstract.tex
\begin{abstract}
Automating optimization modeling from natural language with large language models (LLMs) faces two key challenges. First, training corpora lack structural diversity. Second, data generation pipelines remain static and decoupled from model learning. To address these challenges, we propose \textbf{EvoOptiGraph}, a novel framework where data and model co-evolve, driven by model weaknesses. EvoOptiGraph represents each mixed-integer linear program (MILP) as an attributed bipartite graph and applies validity-preserving evolutionary operators to generate structurally diverse instances. The evolved graphs are converted into solver code and natural language via deterministic compilation and verified back-translation. Training proceeds in two stages: supervised fine-tuning (SFT) on an initial dataset, followed by reinforcement learning with verifiable rewards (RLVR), where graph-derived weakness signals guide the generation of new instances targeting the model's failures. This forms a closed loop that continuously updates the training distribution. Empirical results on six public datasets show that EvoOptiGraph significantly outperforms larger generalist models, agentic methods, and specialized baselines in accuracy, executability, and generalization. These results demonstrate that targeted data–model coevolution is an effective strategy for improving LLMs on optimization modeling tasks.
\end{abstract}

%% file: 01_introduction.tex
\section{Introduction}

Optimization modeling converts a natural-language decision problem into a formal mathematical program that can be solved by an optimization engine. This capability is central to applications in logistics, manufacturing, energy, and finance \citep{singh2012overview, antoniou2007practical}, but in practice it still depends heavily on human expertise. A modeler must identify decision variables, objectives, constraints, and domain assumptions, then translate them into solver-executable code. Automating this process is therefore an appealing but demanding testbed for large language models (LLMs). Recent work has explored prompt engineering \citep{ramamonjison2023nl4opt}, multi-agent reasoning \citep{ahmaditeshnizi2024optimus, xiao2023chain}, and domain-specific fine-tuning \citep{huang2025orlm, jiang2024llmopt} to improve natural-language-to-optimization modeling.

Two challenges limit current progress. First, training data for optimization modeling is narrow relative to the hypothesis space of valid formulations. Benchmarks such as NL4Opt \citep{ramamonjison2023nl4opt} and MAMO \citep{huang2024mamo} are valuable, but they cover only a small subset of the structural patterns encountered in mathematical programming \citep{xiao2025survey}. Existing synthesis pipelines mainly vary coefficients or textual phrasing within fixed templates, which improves scale but not structural breadth. Text-only evolution methods can generate lexical diversity, yet they offer weak control over the underlying optimization structure \citep{li2024towards}. Second, most training pipelines are static: a dataset is constructed once, and the model is then optimized against that frozen distribution. Such a setup makes it difficult to identify which structural regimes the model fails on and to generate new instances that specifically target those weaknesses.

We address these limitations with \textbf{EvoOptiGraph}, a weakness-driven coevolution framework of data and models that couples data generation and model training through an explicit structural representation. Our starting point is to represent each mixed-integer linear program (MILP) as an attributed bipartite graph whose nodes correspond to variables and constraints, and whose edges encode nonzero coefficients. This representation is expressive enough to recover solver-executable code, but structured enough to support validity-preserving genetic operations and fine-grained failure analysis. EvoOptiGraph uses this graph space in two ways: it evolves new optimization instances beyond parameter-level perturbations, and it diagnoses model weaknesses through graph-derived structural features computed on held-out validation problems. The training loop is closed rather than static. We first warm-start the model with supervised fine-tuning (SFT) on an initial population of verified instances. We then estimate a weakness profile for the current model by relating graph features to observed failures. That weakness profile guides further graph evolution, generating instances that are diverse, solvable, and structurally aligned with the model's current blind spots. The newly evolved instances then feed into a reinforcement learning stage with verifiable rewards (RLVR), where rewards are defined by solver executability and solution correctness. In this way, the training distribution adapts as the model changes.

Our contributions are threefold:

\begin{itemize}
    \item \textbf{We propose a novel closed-loop coevolution framework for optimization modeling.} Instead of separating data synthesis from model optimization, EvoOptiGraph links them through graph-level weakness analysis, enabling the training distribution to adapt to the model's evolving failure modes.

   \item \textbf{We design a novel controllable graph-based generator for optimization instances.} Crossover and mutation operate directly on attributed bipartite graphs, enabling structural-level evolution beyond parameter perturbations. Each evolved graph is deterministically compiled into solver code and back-translated into natural language, supporting systematic exploration of diverse optimization structures.

    \item \textbf{We develop an effective weakness-driven model adaptation with verifiable feedback.} After supervised fine-tuning, we use graph-derived weakness signals to evolve new instances that target the model's failures, and optimize the model with RLVR, where rewards are defined by solver executability and solution correctness.
\end{itemize}

Experiments on six public benchmarks, including ComplexOR, MAMO, and NL4Opt, show that our framework improves optimization-modeling performance over strong baselines and yields better generalization to structurally challenging instances. Ablation studies further indicate that both components of the loop matter: graph-based structural evolution expands the support of the training data, while weakness-driven generation makes that expansion relevant to the model's actual errors.

%% file: 02_related_work.tex
\section{Related Work}

\subsection{LLMs for Optimization Modeling}

Recent work studies how LLMs translate natural-language problem statements into mathematical formulations or solver-executable code. Early efforts relied on prompt-based pipelines, such as the NL4Opt benchmark \citep{ramamonjison2023nl4opt}, which highlighted both the promise of LLMs and their tendency to omit constraints or hallucinate assumptions. Subsequent work improves inference-time reasoning via decomposition, multi-agent collaboration, and expert-style prompting, including OptiMUS \citep{ahmaditeshnizi2024optimus}, Chain-of-Experts \citep{xiao2023chain}, and AutoFormulation \citep{astorga2024autoformulation}. In parallel, task-specific models are developed through supervised fine-tuning, such as ORLM \citep{huang2025orlm} and LLMOPT \citep{jiang2024llmopt}.

These approaches improve prompting, reasoning, or task adaptation under a fixed training distribution. In contrast, we focus on constructing and iteratively adapting the data distribution to expose structurally informative instances during training. EvoOptiGraph is complementary to prior modeling architectures and can be combined with stronger inference-time reasoning.

\subsection{Data Synthesis for Optimization Modeling}

Data quality is a key bottleneck for optimization-modeling LLMs \citep{xiao2025survey}. A common approach is parametric template expansion, where canonical problem families are instantiated by varying sizes, coefficients, or templates \citep{lu2025optmath}. While controllable and verifiable, such methods generate instances close to fixed structural templates. Recent work explores broader synthesis: MILP-Evolve \citep{li2024towards} iteratively evolves MILP classes via text-based operators on seed problems, OptMATH \citep{lu2025optmath} constructs aligned triplets via bidirectional generation and solution matching, and MIPLIB-NL \citep{li2026constructing} reverse-generates natural-language descriptions for real-world MILP instances. Other works focus on generating MILP instances primarily for solver evaluation and algorithm stress-testing (e.g., G2MILP~\citep{geng2023deep, alipour2023enhanced}, MILP-StuDio~\citep{liu2024milp}, and instance-space evolution~\citep{smith2015generating, liu2024instance, bowly2019stress}), while our goal is to create training data for LLM-based modeling from natural language.

Our work is closely related to structure-aware instance generation but differs in two aspects. First, EvoOptiGraph applies genetic operations directly on graph representations that preserve optimization structure, rather than relying on textual mutation or parameter perturbation. Second, generation is adaptive: instances are scored by alignment with model weaknesses, enabling dynamic rather than static synthesis.

\subsection{Model Training and Reinforcement Learning}

SFT is the standard starting point, but token-level likelihood does not directly optimize solver correctness or executability. RL with external verification is therefore widely adopted, including outcome-based rewards \citep{chen2025solver}, process supervision with reward models \citep{dai2024process, zhu2025retrieval}, and preference-based objectives such as DPO \citep{zheng2025cold}. Other approaches combine LLMs with search strategies, including beam search, Monte Carlo Tree Search, and evolutionary self-improvement \citep{wang2025bppsearchenhancingtreethought, zheng2025monte, liu2024evolution, ye2024reevo}. For operations research tasks, OR-PRM evaluates intermediate reasoning steps \citep{wangor}. However, these methods typically rely on fixed training datasets and do not adapt data distributions to model weaknesses.

EvoOptiGraph builds on this line of work but shifts the focus to data generation. We also optimize with verifiable rewards, but instead of sampling from a static pool, we diagnose structural error patterns and evolve new instances that target them, turning RL into an adaptive data-model coevolution loop.

%% file: 03_methods.tex
\section{Methodology}

\subsection{Problem Formulation and Graph Representation}

We study \emph{natural-language-to-optimization modeling}: given a problem description $q\in \mathcal{Q}$, where $\mathcal{Q}$ denotes the space of natural-language descriptions, the goal is to generate a correct optimization model together with its solver-executable implementation. We focus on mixed-integer linear programs (MILPs), which cover a large family of practical decision problems. A MILP can be written as
\[
\begin{aligned}
\min/\max \quad & \mathbf{c}^\top \mathbf{x} \\
\text{s.t.} \quad & \mathbf{A} \mathbf{x} \bowtie \mathbf{b}, \\
& \mathbf{l} \leq \mathbf{x} \leq \mathbf{u}, \\
& x_i \in \mathbb{Z} \text{ for } i \in \mathcal{I}, \quad x_j \in \mathbb{R} \text{ for } j \in \mathcal{R},
\end{aligned}
\]
where $\mathbf{x} = (x_1, \dots, x_n)$ is the decision vector, $\mathbf{c}$ is the objective coefficient vector, $\mathbf{A}$ is the constraint matrix, $\mathbf{b}$ is the right-hand-side vector, and $\bowtie \in \{\leq, =, \geq\}^m$ specifies the constraint senses. Variable bounds are denoted by $\mathbf{l}$ and $\mathbf{u}$, and $\mathcal{I}$ and $\mathcal{R}$ index integer and continuous variables, respectively. We denote the algebraic formulation by $\mathcal{P} = (\min/\max, \mathbf{c}, \mathbf{A}, \mathbf{b}, \bowtie, \mathbf{l}, \mathbf{u}, \mathcal{I}, \mathcal{R})$, and its solver-executable program (e.g., Gurobi code) by $\mathcal{V}(\mathcal{P})$.

To support both structured generation and failure analysis, we represent each MILP as an attributed bipartite graph $\mathcal{G} = (V, E, \mathcal{A})$, following the variable--constraint graph representation widely used in learning-based MILP research \citep{gasse2019exact, fan2023smart}. Specifically: (1) Nodes: $V = V_{\text{con}} \cup V_{\text{var}}$, where $V_{\text{con}} = \{s_1, \dots, s_m, s_{\text{obj}}\}$ contains one node per constraint together with a dedicated objective node, and $V_{\text{var}} = \{x_1, \dots, x_n\}$ contains one node per decision variable. (2) Edges: $E \subseteq V_{\text{con}} \times V_{\text{var}}$; an edge $(s_i, x_j)$ is present when variable $x_j$ appears in constraint $s_i$ or in the objective with a nonzero coefficient, and the edge weight stores that coefficient. (3) Attributes: A constraint node stores its type, right-hand side, and relational sense; a variable node stores its lower bound, upper bound, and variable type. This graph serves as an internal representation for data generation and weakness diagnosis. It is \emph{manipulable} (crossover and mutation operate directly on variables, constraints, coefficients, and substructures) and \emph{analyzable} (graph statistics summarize structural regimes where the model succeeds or fails). A valid graph has exactly one objective node, complete variable metadata, and no isolated components; from any valid $\mathcal{G}$ we deterministically recover $\mathcal{P}$ and $\mathcal{V}(\mathcal{P})$.

The learning problem is therefore to estimate a mapping $f_\theta: \mathcal{Q} \rightarrow \mathcal{V}$ from natural-language descriptions to solver-executable programs. EvoOptiGraph improves this mapping indirectly by controlling the structure of the training data and adapting that structure over time.

\begin{figure}[t]
    \centering
    \includegraphics[width=1\linewidth]{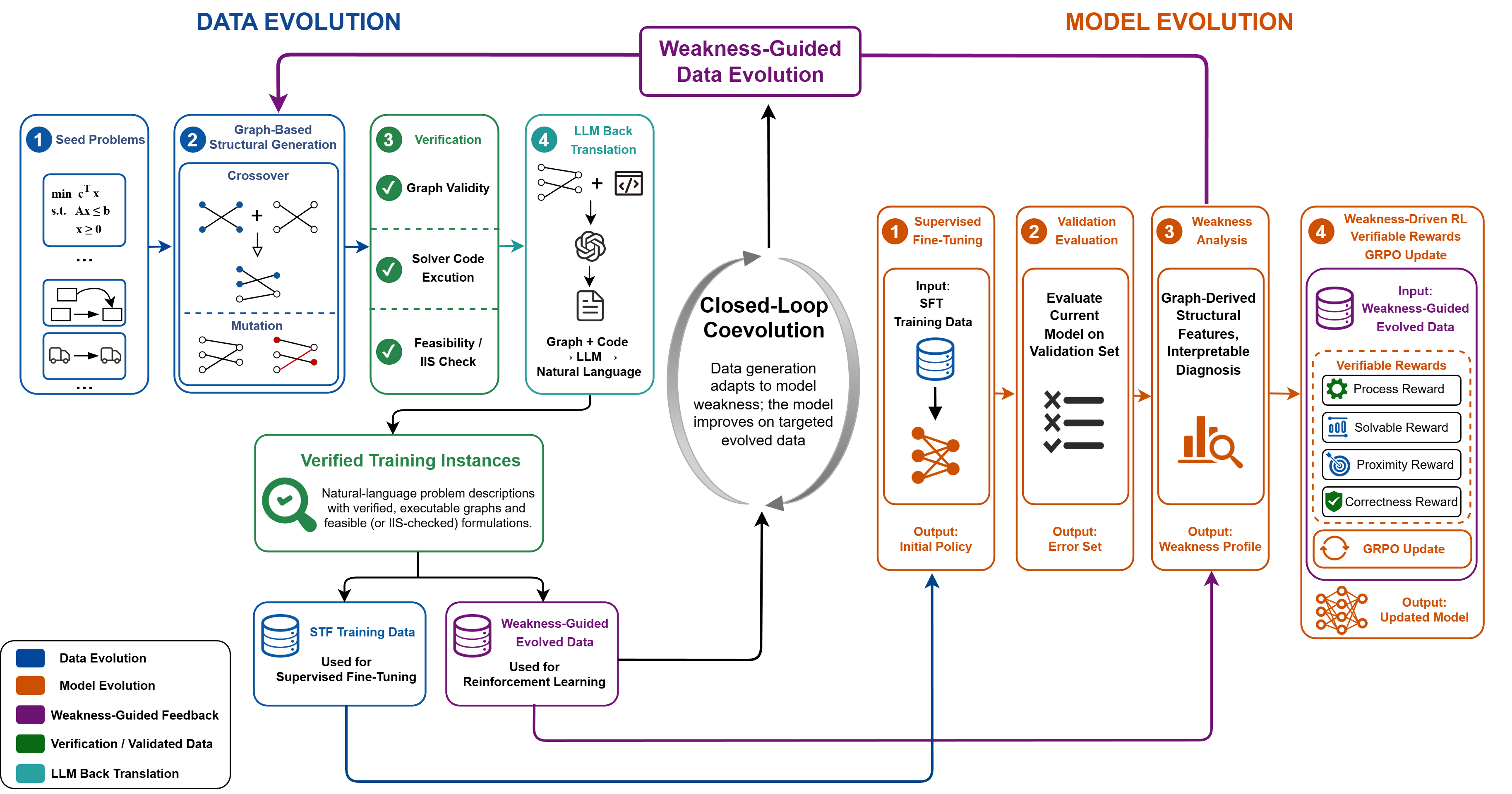}
    \caption{Weakness-Guided Data–Model Coevolution Framework.}
    \vspace{-15pt}
    \label{fig:overall}
\end{figure}

\subsection{Controllable Data Generation via Graph Evolution}
\label{sec:data-evolution}

We generate training instances by evolving graphs rather than only perturbing textual descriptions or numeric parameters. The initial population is built from 53 expert-designed seed generators from OptMATH \citep{lu2025optmath}, each corresponding to a classical problem family such as transportation, knapsack, or facility location. Sampling the parameters of these generators yields an initial dataset $\mathcal{D}_0 = \{(q_i, \mathcal{G}_i, c_i)\}_{i=1}^{N_0}$, where $q_i$ is a natural-language description, $\mathcal{G}_i$ is the graph representation of the optimization instance, and $c_i = \mathcal{V}(\mathcal{P}_i)$ is the executable solver program compiled from the graph. 

\paragraph{Genetic operations with structural constraints.}
We apply genetic operators directly to $\mathcal{G}$ subject to validity constraints. \textbf{Crossover} exchanges compatible subgraphs between two parent instances. In the default setting, crossover is performed within the same problem class to preserve semantic coherence; a smaller number of cross-class operations is allowed to increase structural diversity. The objective node is handled separately so that each offspring still contains exactly one objective. \textbf{Mutation} acts at three levels. \emph{Parameter mutation} perturbs numerical quantities such as coefficients, bounds, and right-hand sides. \emph{Structural mutation} edits local graph topology by inserting, deleting, or modifying variables or constraints while preserving linearity and metadata consistency. \emph{Scale mutation} changes problem size, e.g., expanding supply/demand nodes in a transportation instance. Every offspring is filtered by deterministic validity checks before entering the candidate pool.

\begin{figure}[htbp]
\centering
\includegraphics[width=1\linewidth]{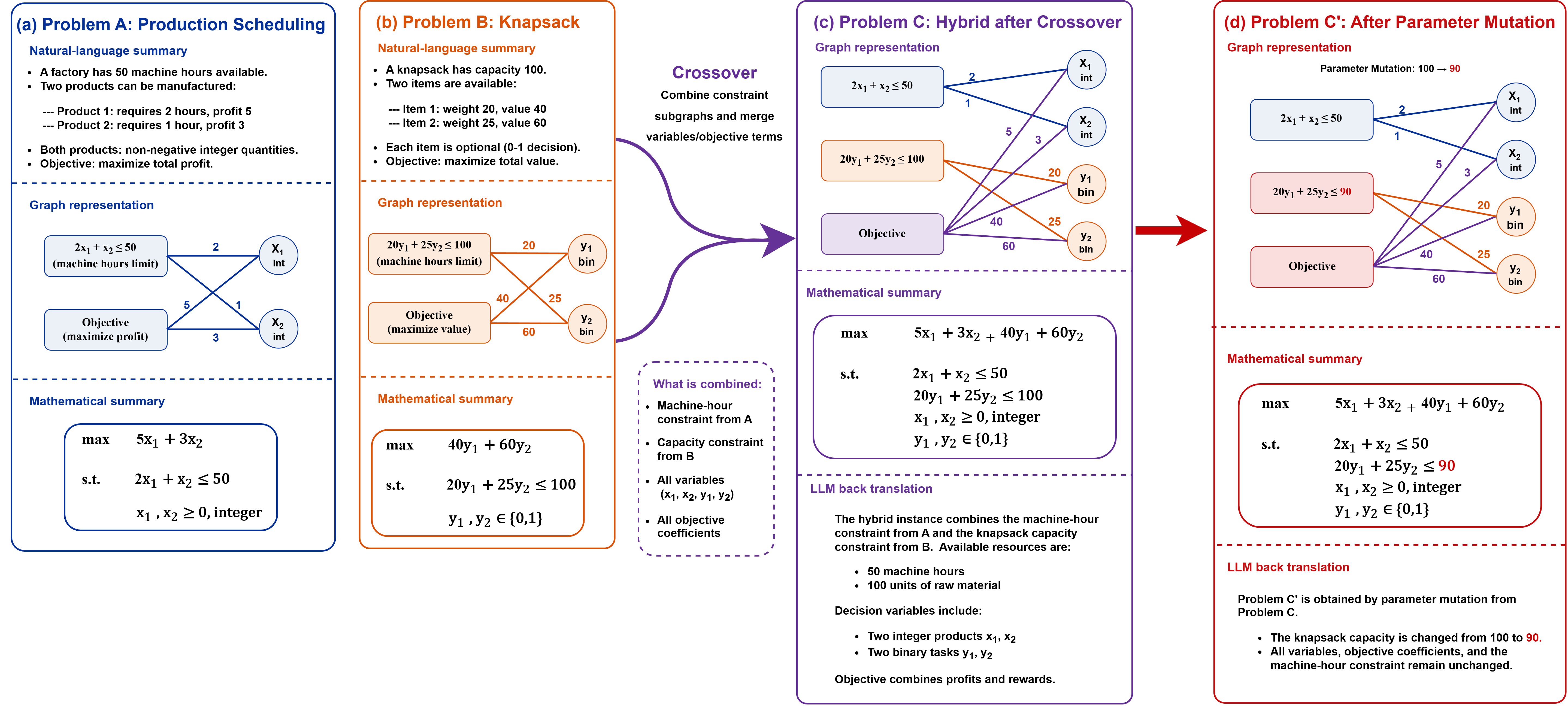}
\caption{Crossover and mutation example. (a) Production scheduling; (b) knapsack; (c) hybrid after crossover (C); (d) mutated from C (C', capacity 100$\rightarrow$90). Red highlight indicates mutation.}
\vspace{-15pt}
\label{fig:crossover_mutation}
\end{figure}

Figure~\ref{fig:crossover_mutation} shows an example: crossing a production scheduling problem (A) with a knapsack (B) yields hybrid (C); mutating (C)'s knapsack capacity (100→90) gives (C'). See appendix for full details.


\paragraph{Quality control and fitness function.}
Each candidate instance is verified before entering the training pool. The first stage checks graph structural validity. The second stage compiles the graph into Gurobi code and executes it. For feasible instances, we record solver statistics such as solve time and branch-and-bound nodes as difficulty metadata. For infeasible instances, we compute an Irreducible Infeasible Subset (IIS) \citep{gurobi,chinneck2008feasibility}; instances with non-empty IIS are retained because they still define meaningful optimization structures, while candidates that fail to compile or execute are discarded. Approximately 3\% infeasible examples are retained only in SFT and are excluded from weakness analysis and RL, so RL examples retain defined ground-truth objectives. Structural validity and solver feasibility rates are reported in Appendix~\ref{app:validity}. Verification metadata is cached to avoid redundant solver calls. The fitness function balances three normalized objectives (min-max scaling to $[0,1]$):
\[
f(\mathcal{G}) = \beta \tilde{f}_{\text{diff}} + \gamma \tilde{f}_{\text{div}} + \delta \tilde{f}_{\text{weak}}.
\]
Difficulty $\tilde{f}_{\text{diff}}$ is derived from solver statistics and favors nontrivial instances. Diversity $\tilde{f}_{\text{div}}$ is the average graph edit distance (GED) \citep{gao2010survey, gasse2019exact, fan2023smart} to the current population, discouraging collapse to a few templates. Weakness alignment $\tilde{f}_{\text{weak}}$ is the cosine similarity between the candidate's structural feature vector $\phi(\mathcal{G})$ and the current model weakness vector $\mathbf{w}$ (Section~\ref{sec:weakness}). This term is inactive initially ($\delta=0$) and becomes important once the model has enough validation evidence to estimate its failure modes.

\paragraph{Back translation to natural language.}
For each verified graph, we prompt an LLM agent with its serialization and provenance to generate a natural-language description. A three-stage verification ensures quality: (1) rule-based checks (numerical consistency, coverage of variables and constraints, fidelity to objective and operator types); (2) the agent checks logical compatibility, regenerating up to a fixed budget if needed; (3) periodic value-matching audits on sampled instances (reconstruct solver code via a separate forward-modeling system and compare outcomes). Filter rates and the residual mismatch after filtering are reported in Appendix~\ref{app:audit}.
\subsection{Two-Stage Model Training and Coevolution Loop}
\label{sec:training}

The training loop alternates between model fitting and data evolution: the current model is evaluated on a held-out validation set, and its failure patterns are summarized as a weakness vector $\mathbf{w}$, which biases the generator toward informative instances for RL. Algorithm~\ref{alg:evooptigraph} outlines the procedure.

\begin{algorithm}
\caption{EvoOptiGraph Training Loop}
\label{alg:evooptigraph}
\begin{algorithmic}[1]
\State \textbf{Input:} Seed generators $\mathcal{S}$, base LLM with parameters $\theta_0$, validation set $\mathcal{V}_{\text{val}} = \{(\mathcal{G}_i, y_i)\}_{i=1}^{N_{\text{val}}}$, number of iterations $T$
\State \textbf{Output:} Optimized model parameters $\theta_T$

\State $\mathcal{D}_0 \leftarrow \text{EvoGenerate}(\mathcal{S})$ \hfill $\triangleright$ initial population 
\State $\theta_{\text{warm}} \leftarrow \arg\min_{\theta} \mathcal{L}_{\text{SFT}}(\theta; \mathcal{D}_0)$ \hfill $\triangleright$ warm-up SFT
\State $\theta_0 \leftarrow \theta_{\text{warm}}$

\For{$t = 1$ to $T$}
    \State $\mathcal{E}_t \leftarrow \{ (\mathcal{G}, y) \in \mathcal{V}_{\text{val}} \mid M_{\theta_{t-1}}(\mathcal{G}) \neq y \}$ \hfill $\triangleright$ error set
    \State $\Phi_t \leftarrow \{ \phi(\mathcal{G}) \mid \mathcal{G} \in \mathcal{V}_{\text{val}} \}$ \hfill $\triangleright$ feature vectors
    \State $\mathbf{w}_t \leftarrow \text{SHAP}\big( \text{XGBoost}(\{(\phi(\mathcal{G}), \mathbf{1}_{(\mathcal{G},y)\in\mathcal{E}_t})\}) \big)$
    \State Evolve new data using fitness: $f(\mathcal{G}) = \beta\tilde{f}_{\text{diff}} + \gamma\tilde{f}_{\text{div}} + \delta \tilde{f}_{\text{weak}}$
    \State  $\mathcal{D}_{\text{RL}} \gets \text{EvoGenerate}(\mathcal{S}, f(\mathcal{G}), \mathbf{w}_t)$ \hfill $\triangleright$ weakness-targeted data
    
    \State 
    Sample candidate responses $\{y_i\}_{i=1}^{G} \sim \pi_{\theta_{t-1}}(\cdot \mid q)$ for each $(q,f_{\text{opt}}) \in \mathcal{D}_{\text{RL}}$
    \State Extract code blocks $\{c_i\}_{i=1}^{G}$ from $\{y_i\}_{i=1}^{G}$ and execute them in parallel
    \State Compute staged rewards $\{R_i\}_{i=1}^{G}$ using $R_{\text{process}} + R_{\text{solvable}} + R_{\text{proximity}} + R_{\text{correctness}}$
    \State $\theta_t \gets \arg\max_{\theta} J_{\text{GRPO}}(\theta; \theta_{t-1}, \mathcal{D}_{\text{RL}}, \{R_i\})$ \hfill $\triangleright$ RL update
\EndFor

\State \textbf{return} $\theta_T$
\end{algorithmic}
\end{algorithm}

\paragraph{Warm-up supervised fine-tuning.}


We train a base LLM on the initial dataset $\mathcal{D}_0$ by standard next-token prediction: 
$\mathcal{L}_{\text{SFT}} = -\sum_t \log P_\theta(c_t \mid c_{<t}, q)$, 
where $\mathcal{D}_0$ is generated from seed problems via graph evolution, 
$q$ is the problem description and $c_t$ is the $t$-th token of the solver code, 
yielding an initial policy $M_{\text{warm}}$ competent for error analysis.

\paragraph{Structured weakness analysis.}
\label{sec:weakness}
In each coevolution round, we evaluate the current model on a held-out validation set whose instances are paired with graphs and ground-truth outcomes. For every graph $\mathcal{G}$, we compute a structural feature vector $\phi(\mathcal{G})$ that includes size statistics (numbers of variables and constraints), type information (e.g., proportions of binary and integer variables), sparsity and density measures, graph-topological features (such as average degree and spectral quantities), and coefficient statistics (see Appendix for a detailed list). These features provide a low-dimensional summary of the optimization. We train an XGBoost classifier \citep{chen2016xgboost} to predict whether the model solves an instance correctly from $\phi(\mathcal{G})$. To extract interpretable feature contributions, we compute SHAP values \citep{lundberg2017unified} on the error cases. The weakness weight for feature $k$ is $w_k = \frac{1}{|\mathcal{E}|} \sum_{i \in \mathcal{E}} \text{SHAP}_{i,k}$, where $\mathcal{E}$ is the set of validation instances answered incorrectly by the current model. The resulting vector $\mathbf{w} = [w_1, \dots, w_d]$ summarizes structural properties most associated with failure and serves as a practical prioritization signal, not a causal statement.

\paragraph{Weakness-driven reinforcement learning.}
Using $\mathbf{w}$, we rerun the evolutionary generator and activate the weakness-alignment term $\tilde{f}_{\text{weak}} = \text{cosine\_similarity}(\phi(\mathcal{G}), \mathbf{w})$, which biases evolution toward instances whose structure resembles the model's current failure regime. We then generate a new dataset $\mathcal{D}_{\text{RL}}$ by evolving instances with this weakness-aligned fitness for RL. Appendix~\ref{app:weakness-guidance} details how $\mathbf{w}_t$ changes which candidates survive tournament selection.

We formulate the update as reinforcement learning with verifiable rewards (RLVR) \citep{lambert2024tulu, ye2025vla, jiang2026verifiable}. Each training example comprises a description $q$ and its ground-truth optimal objective value $f_{\text{opt}}$; the corresponding graph and solver code are used offline. During RL, the model generates a response, we extract and execute its code with Gurobi. The reward is a multi-stage sum:
\[
R(c) = R_{\text{process}}(c) + R_{\text{solvable}}(c) + R_{\text{proximity}}(c) + R_{\text{correctness}}(c),
\]
where $R_{\text{process}}(c)=0.2$ if code executes without Python error, else $0$. Let $\mathrm{relerr} := \frac{|f_{\text{model}} - f_{\text{opt}}|}{|f_{\text{opt}}| + \delta}$ denote the relative error between the solver's returned objective $f_{\text{model}}$ and the ground-truth objective value $f_{\text{opt}}$, with $\delta = 10^{-8}$. If the solver returns a feasible solution, we set $R_{\text{solvable}}=0.1$ and $R_{\text{proximity}}=0.2\cdot\max(0,1-\mathrm{relerr}/0.1)$; otherwise both are $0$. Thus $R_{\text{proximity}}$ decays linearly from $0.2$ at $\mathrm{relerr}=0$ to $0$ at $\mathrm{relerr}=0.1$ (10\% relative error). Intuitively, a solution whose relative error is within 10\% of the ground truth is more likely to capture the main structure (e.g., minor coefficient errors), while a large deviation tends to signal a fundamental mistake. The correctness bonus is $R_{\text{correctness}}=0.5$ when solver status is \textit{optimal} and $\mathrm{relerr}\le\tau_{\text{rel}}$ ($\tau_{\text{rel}}=10^{-4}$), else $0$. Total reward in $[0,1]$. This layered design encourages runnable code, any feasible solution, numerical closeness (within 10\%), and exact matches.

To optimize the policy $\pi_\theta$, we adopt group-relative policy optimization (GRPO) \citep{shao2024deepseekmath, guo2025deepseek}. For each prompt $q$, we sample $N$ candidate responses $\{y_i\}_{i=1}^N$ from the current policy $\pi_{\theta_{\text{old}}}$, evaluate their rewards $\{R_i\}_{i=1}^N$, and normalize them within the group to obtain advantages $\hat{A}_i = (R_i - \mu_R) / \sigma_R$. We then maximize
{\small
\[
J_{\text{GRPO}}(\theta) = \mathbb{E}_{q \sim \mathcal{D}_{\text{RL}},\, \{y_i\} \sim \pi_{\theta_{\text{old}}}} \left[ \frac{1}{N} \sum_{i=1}^N \left( \min\left( r_i(\theta) \hat{A}_i,\ \text{clip}(r_i(\theta), 1-\epsilon, 1+\epsilon) \hat{A}_i \right) - \widetilde{\beta} \cdot \mathbb{D}_{\text{KL}}(\pi_\theta \| \pi_{\text{ref}}) \right) \right],
\]}
where $r_i(\theta) = \frac{\pi_\theta(y_i \mid q)}{\pi_{\theta_{\text{old}}}(y_i \mid q)}$ is the importance ratio, $\epsilon$ is the clipping hyperparameter, $\widetilde{\beta}$ controls the KL penalty strength, and $\pi_{\text{ref}}$ is the SFT reference policy. The KL term regularizes the RL update so that the policy does not drift too far from the supervised initialization while still exploiting solver-based rewards.

\paragraph{Iterative coevolution loop.}
At each iteration $t$, we first diagnose the current model $M_t$ on the validation set to obtain the weakness vector $\mathbf{w}_t$. We then evolve new instances using weakness-aligned fitness to generate $\mathcal{D}_{\text{RL}}$, and update $M_t$ to $M_{t+1}$ via GRPO. This closed loop makes the data distribution co-evolve with the model, generating new data in response to observed errors rather than statically. Figure \ref{fig:overall} shows our overall framework diagram.

%% file: 04_experiments.tex
\input{table/table1}
\section{Experiments}

\subsection{Experiment Setup}
\textbf{Data Generation.} We generate training data via genetic graph evolution. Starting from 53 expert-designed seed problem families \cite{lu2025optmath}, the generator applies mutation and crossover operators to produce new optimization instances. Beyond the original 53 families, we additionally construct 30 cross-family hybrid categories through inter-class crossover. These hybrid categories make up about 20\% of the data generated in each round.
In terms of scale, we use 5,000 generated instances for the initial SFT stage. After warm-up, each RL iteration uses a newly generated pool of 1,000 instances. We further generate 2,000 additional instances per iteration for structured weakness analysis, which are used to train the XGBoost classifier.

\textbf{Benchmarks.} Our evaluation is conducted on six public benchmarks commonly used in the OR community: NL4Opt \cite{ramamonjison2023nl4opt}, MAMO, which is divided into EasyLP and ComplexLP \cite{huang2024mamo}, NLP4LP \cite{ahmaditeshnizi2024optimus}, ComplexOR \cite{xiao2023chain}, IndustryOR \cite{huang2025orlm}.
See Appendix for more details about these benchmarks.



\textbf{Baseline.} To ensure a thorough comparison, we benchmark against a comprehensive set of baselines spanning three categories: zero-shot generalist LLMs, specialized fine-tuned LLMs, and agentic methods. The zero-shot generalist LLMs include GPT-4o \cite{openai2026gpt4o}, DeepSeek-V3 \cite{liu2024deepseek}, Qwen3-32B \cite{yang2025qwen3}, and Qwen2.5-72B-Instruct \cite{qwen2025qwen25technicalreport}. The specialized fine-tuned LLMs include ORLM (8B) \cite{huang2025orlm}, LLMOPT (14B) \cite{jiang2024llmopt}, and OptMATH (32B) \cite{lu2025optmath}. The agentic methods include multi-step reasoning frameworks that orchestrate LLMs to solve problems, such as OptiMUS-v0.3 \cite{ahmaditeshnizi2024optimus}, Chain-of-Thought (CoT) \cite{wei2022chain} and Chain-of-Experts (CoE) \cite{xiao2023chain}. For OptiMUS-v0.3, we use the results reported by \cite{zhang2026sacoptsemanticanchorsiterative}, while for CoE, CoT, and ORLM, we use the results reported by \cite{xiao2025survey}, as their evaluations are conducted on the same dataset as ours, namely the cleaned version released by \cite{xiao2025survey}. For OptMATH and LLMOPT, marked with * in Table~\ref{tab:main}, we report reference results on the original uncleaned splits. Exact sources and the single-shot protocol used for LLMOPT are given in Appendix~\ref{app:provenance}.

\textbf{Implementation Details.} In the back-translation stage of data generation, we use DeepSeek-V3 \cite{liu2024deepseek} to generate natural-language descriptions from verified graph instances. We adopt the prompting framework of OptMATH \cite{lu2025optmath}, which consists of three stages—generate, check, and regenerate—with full prompt details provided in the appendix. We develop EvoOptiGraph by fine-tuning the publicly available Qwen3-8B \cite{yang2025qwen3} model. Detailed configurations and procedures of the fine-tuning process are also deferred to the appendix. As the primary evaluation metric, we adopt pass@1 accuracy, which measures whether the optimal objective value obtained from the generated optimization code matches the reference value provided by the benchmark. 

\subsection{Main Results}
Table \ref{tab:main} presents the overall Pass@1 results on six OR benchmarks. Among methods evaluated on the cleaned splits, EvoOptiGraph achieves the best macro-average Pass@1, 78.1\%. Mean and standard deviation over 2 training seeds and 3 evaluation runs are reported in Appendix~\ref{app:mean-std}. It consistently outperforms zero-shot LLMs, agentic methods, and prior fine-tuned models, showing the effectiveness of our two-stage model training and coevolution loop. Across individual benchmarks, EvoOptiGraph achieves the strongest or tied-best results on nearly all tasks, demonstrating robust generalization across diverse optimization problem types. Notably, this performance is achieved with an 8B backbone, yet it still surpasses substantially larger models. This suggests that, for optimization-oriented code generation, targeted data evolution and structured weakness adaptation can be more effective than simply increasing model scale.
\subsection{Data diversity comparison}
\label{sec:diversity}
We evaluate data diversity using the Structural Similarity metric introduced in GenBench-MILP \cite{luo2026generatedinstancestrulyuseful}. Specifically, we compare the similarity between the generated instances and the 53 original problem types for two data generation methods: OptMATH \cite{lu2025optmath} and our method. Under this metric, lower similarity indicates higher diversity. For each method, we first generate 5,000 instances in total. Then, to ensure a statistically meaningful comparison, we randomly sample 212 instances at a time, compute their structural similarity to the 53 original problem types, and repeat this process 50 times. We report the average similarity over these 50 runs. The results show that the instances generated by our method achieve a structural similarity of 66.93\%, whereas those generated by OptMATH achieve 93.04\%. This substantial gap indicates that our method produces significantly more diverse data. A possible reason is that OptMATH mainly perturbs the explicit parameters of the original problems, while our method modifies the underlying problem structure itself, leading to richer structural variations.
\begin{figure}[t]
    \centering
    \includegraphics[width=1\linewidth]{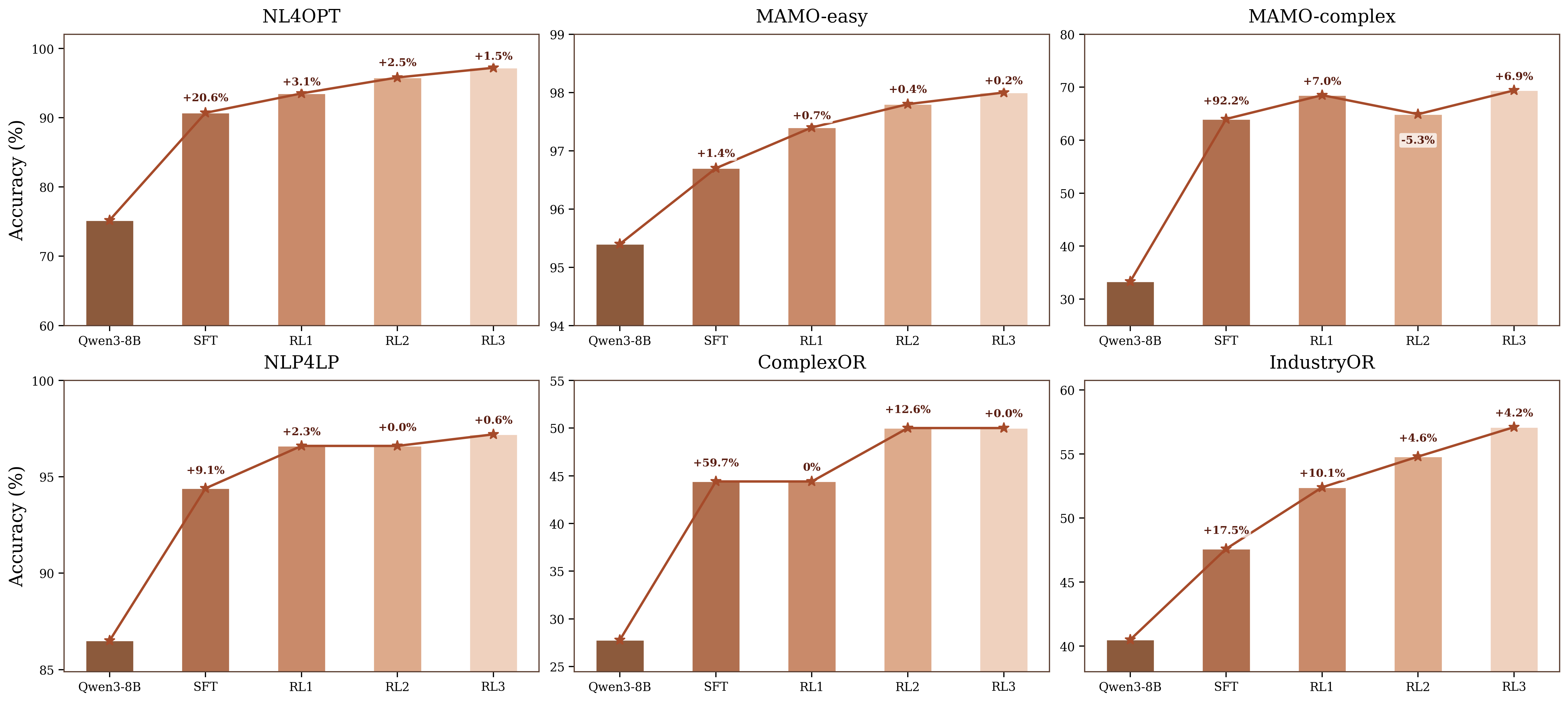}
    \caption{Accuracy trajectory across the coevolution loop. The relative improvement of current iteration over the previous one is demonstrated on the corresponding bar.}
    \vspace{-17pt}
    \label{fig:loop}
\end{figure}
\subsection{Analysis of Coevolution Loop}
We analyze the coevolution loop by tracking model performance from the base model to the warm-up SFT stage and three subsequent reinforcement-learning rounds. Figure~\ref{fig:loop} reveals three observations. First, the warm-up SFT stage provides the largest foundational improvement across all benchmarks, especially on structurally challenging datasets such as MAMO-complex and ComplexOR, indicating that supervised training on verified optimization data is critical for establishing basic optimization modeling and code-generation competence. Second, the subsequent coevolution rounds consistently build on top of the SFT model on most benchmarks, yielding steady additional gains. This trend suggests that the weakness-guided generate-and-train loop continues to uncover informative failure modes even after supervised learning has saturated, rather than merely repeating the same data distribution.  We also observe a non-monotonic trajectory on MAMO-complex, where performance drops at RL2 before recovering at RL3, which suggests that adaptive data evolution may temporarily overspecialize to certain failure regimes but remains self-correcting over subsequent rounds. 
\input{table/table2}
\subsection{Ablation Study}
To investigate the effectiveness of the two-stage training scheme and the coevolution loop, we conduct an ablation study with two variants: SFT-only and RL-only. Compared with SFT-only and RL-only, EvoOptiGraph improves the average accuracy by 9.6\% and 14.7\%, respectively. Detailed results are presented in the last row of Table \ref{tab:main} and Table \ref{ablation}. We find that SFT-only brings clear gains in the first round, but its improvement soon saturates and can even deteriorate in later rounds. One possible explanation is that repeated supervised updates on synthetic data may cause the model to overfit to recurring data patterns, while failing to further improve outcome-level optimization correctness. In contrast, RL-only shows only limited improvement, likely because without an SFT warm-up, the model does not have a sufficiently strong initialization, making reward-driven learning unstable and inefficient. For the accuracy trajectory, please refer to Figures \ref{fig:RL} and \ref{fig:STF} in the appendix.

To isolate the contribution of weakness-driven targeting, we compare three instance-selection strategies under the same SFT warm-up, GRPO recipe, and per-round data budget (Table~\ref{tab:fitness_ablation}). Full fitness uses weakness alignment, with $\delta$ following the schedule in Appendix~\ref{app:ga}. The $\delta=0$ variant keeps only the difficulty and diversity terms. Uniform sampling draws evolved instances with equal probability. Enabling weakness alignment improves the macro-average Pass@1 by 10.8\% over $\delta=0$ (78.1\% vs.\ 67.3\%). Graph-based evolution with difficulty and diversity filtering still outperforms uniform sampling by 2.5\% (67.3\% vs.\ 64.8\%). Structural evolution is therefore beneficial on its own, and aligning new instances with the model's failure distribution supplies a further, larger gain.
\input{table/table4}

%% file: table/table1.tex
\definecolor{mypink}{RGB}{252, 218, 218}

\begin{table}[t]
\caption{The overall performance of \texttt{EvoOptiGraph} and baselines with Pass@1 accuracy (\%) on six OR benchmarks. Instance counts are NL4OPT 214, MAMO-EasyLP 545, MAMO-ComplexLP 111, NLP4LP 178, ComplexOR 18, and IndustryOR 42. Scores marked with * are reference-only results on uncleaned splits (Appendix~\ref{app:provenance}) and are excluded from the ranking; missing entries are denoted with (-). Among cleaned-split methods, the best result in each column is \textbf{bold} and the second-best is \underline{underlined}; tied values share the same mark.}

\label{tab:main}
\centering
\resizebox{\textwidth}{!}{%
\begin{tabular}{lcccccccc}
\toprule
\multirow{2}{*}{Model} & \multirow{2}{*}{Params} & \multirow{2}{*}{NL4OPT} & \multicolumn{2}{c}{MAMO} & \multirow{2}{*}{NLP4LP} & \multirow{2}{*}{CompOR} & \multirow{2}{*}{IndOR} & \multirow{2}{*}{Avg.} \\
\cmidrule(lr){4-5}
 & & & EasyLP & ComplexLP & & & & \\
\midrule
\multicolumn{9}{c}{\textbf{Zero-shot LLMs}} \\
\midrule
\texttt{GPT-4o}            & Closed & 73.4             & 95.2             & 46.0             & 87.6             & 38.9             & \underline{66.7} & 68.0 \\
\texttt{DeepSeek-V3}       & 671B   & \underline{79.9} & 96.7             & 55.0             & \underline{94.9} & 44.4             & \textbf{69.1}    & \underline{73.3} \\
\texttt{Qwen3-32B}         & 32B    & 76.1             & \underline{96.9} & 32.4             & 90.5             & 38.9             & 57.1             & 65.3 \\
\texttt{Qwen2.5-72B-Inst}  & 72B    & \underline{79.9} & 95.8             & 33.3             & 89.9             & 38.9             & 61.9             & 66.6 \\
\midrule
\multicolumn{9}{c}{\textbf{Agentic Methods}} \\
\midrule
\texttt{OptiMUS-v0.3}      & Closed & 79.8             & 92.4             & 52.1             & 89.8             & \underline{52.2} & 54.3             & 70.1 \\
\texttt{CoT}               & Closed & 62.2             & 49.5             & 42.3             & 74.7             & 39.2             & 40.5             & 51.4 \\
\texttt{CoE}               & Closed & 66.7             & 94.4             & 50.6             & 87.4             & \textbf{57.1}    & 31.2             & 64.6 \\
\midrule
\multicolumn{9}{c}{\textbf{Fine-tuned LLMs}} \\
\midrule
\texttt{ORLM}              & 8B     & 73.8             & 90.4             & \underline{59.5} & 76.4             & 50.0             & 42.9             & 65.5 \\
\texttt{LLMOPT} (origin)   & 14B    & 80.3*            & 89.5*            & 44.1*            & 73.4*            & 35.3*            & 29.0*            & 58.6* \\
\texttt{OptMATH} (origin)  & 32B    & 95.9*            & 89.9*            & 54.1*            & -                & -                & -                & - \\
\rowcolor{mypink}
\texttt{EvoOptiGraph}      & 8B     & \textbf{96.7}    & \textbf{98.0}    & \textbf{69.4}    & \textbf{97.2}    & 50.0             & 57.1             & \textbf{78.1} \\
\bottomrule
\end{tabular}%
}

\vspace{2pt}
{\footnotesize \textbf{Abbreviations:} CompOR: ComplexOR, IndOR: IndustryOR, Avg: Macro-Average, Inst: Instruct.}
\vspace{-10pt}
\end{table}

%% file: table/table2.tex
\begin{table}[t]
\caption{Ablation study with Pass@1 accuracy (\%) on six OR benchmarks. }
\label{ablation}
\centering
\small
\setlength{\tabcolsep}{4pt}
\renewcommand{\arraystretch}{0.9}
\begin{tabular}{lccccccc}
\toprule
\multirow{2}{*}{Type} & \multirow{2}{*}{NL4OPT} & \multicolumn{2}{c}{MAMO} & \multirow{2}{*}{NLP4LP} & \multirow{2}{*}{CompOR} & \multirow{2}{*}{IndOR} & \multirow{2}{*}{Avg.} \\
\cmidrule(lr){3-4}
 & & EasyLP & ComplexLP & & & & \\
\midrule
\texttt{All-SFT} & 85.5 & 96.0 & 55.9 & 91.6 & 38.9 & 42.9 & 68.5 ($\downarrow$ \textbf{9.6}) \\
\texttt{All-RL}  & 72.0 & 96.3 & 33.3 & 87.6 & 38.9 & 52.4 & 63.4 ($\downarrow$ \textbf{14.7}) \\
\bottomrule
\end{tabular}
\vspace{-15pt}
\end{table}



%% file: table/table4.tex
\begin{table}[t]
\caption{Ablation of weakness alignment with Pass@1 accuracy (\%) on six OR benchmarks.}
\label{tab:fitness_ablation}
\centering
\small
\setlength{\tabcolsep}{3.5pt}
\renewcommand{\arraystretch}{0.9}
\begin{tabular}{lccccccc}
\toprule
\multirow{2}{*}{Setting} & \multirow{2}{*}{NL4OPT} & \multicolumn{2}{c}{MAMO} & \multirow{2}{*}{NLP4LP} & \multirow{2}{*}{CompOR} & \multirow{2}{*}{IndOR} & \multirow{2}{*}{Avg.} \\
\cmidrule(lr){3-4}
 & & EasyLP & ComplexLP & & & & \\
\midrule
(a) Full fitness & 96.7 & 98.0 & 69.4 & 97.2 & 50.0 & 57.1 & 78.1 \\
(b) $\delta=0$     & 87.4 & 95.4 & 45.0 & 91.6 & 38.9 & 45.2 & 67.3 \\
(c) Uniform      & 83.2 & 94.7 & 41.4 & 89.9 & 38.9 & 40.4 & 64.8 \\
\bottomrule
\end{tabular}
\vspace{-12pt}
\end{table}

%% file: 05_conclusion.tex
\section{Conclusion and Future Work}

We presented EvoOptiGraph, a weakness-driven coevolution framework of data and models for natural-language-to-optimization modeling. By representing MILPs as attributed bipartite graphs, it unifies structural data generation, failure diagnosis, and model adaptation in a closed loop, combining supervised fine-tuning with reinforcement learning under verifiable solver-based rewards. Experiments on multiple public benchmarks show consistent gains over baselines; our 8B model achieves the best macro-average Pass@1, demonstrating that targeted structural data evolution can be more effective than scaling model size alone. Stage-wise and ablation results confirm that weakness-driven coevolution yields additional gains beyond SFT or RL alone.

Several limitations remain: the current framework is limited to MILPs and the coverage of available seed generators, verification becomes expensive for larger instances, and the weakness vector is a practical prioritization signal rather than a causal account. Future work includes extending EvoOptiGraph to broader optimization classes, improving efficiency, and studying transferability of weakness signals across problem families. We hope this work offers a useful step toward adaptive training pipelines for formal reasoning and optimization with large language models.

%% file: 07_appendix.tex
\section{Benchmarks}
A large number of datasets and dataset-construction methods have been introduced to improve the capability of LLMs in solving OR problems, and these datasets have become increasingly complex and realistic \cite{ahmaditeshnizi2024optimus}. \cite{xiao2025survey} provide a recent and comprehensive survey of benchmark datasets for natural-language-described optimization problems. Their study highlights several recurring issues in existing benchmarks, such as logical inconsistencies, ambiguous problem statements, and incorrect ground-truth answers, and further rectifies part of these errors. 

\paragraph{NL4OPT}

The NL4Opt dataset, introduced by \cite{ramamonjison2023nl4opt} for the NeurIPS 2022 NL4Opt competition, is a benchmark containing approximately 1,100 annotated linear programming problems. It is designed to study the mapping from natural language descriptions to formal optimization models. The original release comprises 713 training instances, 99 validation instances, and 289 test instances. Due to the noise and errors in the original dataset, we follow \cite{xiao2025survey} and use their cleaned version instead. Specifically, our experiments are conducted on 214 cleaned instances.

\paragraph{MAMO}
The MAMO dataset was introduced by  \cite{huang2024mamo} as a benchmark for evaluating large language models on mathematical modeling tasks. It focuses on LP and MILP problems, with an emphasis on the formulation process rather than the solving step itself. The original dataset is divided into two subsets: Easy LP with 652 instances and Complex LP with 211 instances. In our work, rather than using the original release directly, we adopt the cleaned version provided by Xiao et al. After data cleaning, we use 545 easy instances and 111 complex instances in our experiments.

\paragraph{NLP4LP}
The NLP4LP benchmark was introduced by  \cite{ahmaditeshnizi2024optimus} for evaluating the ability of large language models to translate natural language operations research problems into solver-ready code and mathematical formulations. It consists of 269 human-authored LP and MILP problems covering classical operations research domains such as scheduling, knapsack allocation, and production planning. In our work, rather than using the original release directly, we adopt the cleaned version provided by Xiao et al. After data cleaning, we use 178 instances in our experiments.

\paragraph{COMPLEXOR}
The ComplexOR dataset was introduced by \cite{xiao2023chain} as a benchmark for evaluating large language models on challenging operations research modeling tasks. It consists of complex optimization problems drawn from advanced OR case studies and difficult problem classes, such as multi-step lot-sizing, intricate scheduling, and supply chain design. These instances are primarily large-scale or conceptually complex mixed-integer linear programming problems that require multi-step reasoning. In our work, we use 18 instances from this dataset in our experiments.
\paragraph{INDUSTRYOR}
The IndustryOR dataset was introduced by \cite{huang2025orlm} as a benchmark for evaluating large language models on real-world operations research problems. It consists of 100 optimization instances drawn from practical domains such as manufacturing, logistics, finance, and energy. The dataset covers multiple problem categories, including linear, integer, mixed-integer, nonlinear, and other types. In our work, rather than using the original release directly, we adopt the cleaned version provided by Xiao et al. After data cleaning, we use 42 instances in our experiments.

\subsection{Baseline Provenance}
\label{app:provenance}
Table~\ref{tab:main} separates results on the cleaned splits from reference numbers on the original releases. Zero-shot LLMs, OptiMUS-v0.3, CoT, CoE, and ORLM are compared on the cleaned splits of \cite{xiao2025survey}, the same splits used for EvoOptiGraph, under a single-shot Pass@1 protocol with Gurobi verification. CoT, CoE, and ORLM follow the numbers reported by \cite{xiao2025survey}. OptiMUS-v0.3 follows \cite{zhang2026sacoptsemanticanchorsiterative}. OptMATH \cite{lu2025optmath} and LLMOPT \cite{jiang2024llmopt}, marked with * in the table, were evaluated on the original uncleaned splits and are reported only as reference; they are not directly comparable to the cleaned-split results. Because those papers do not report matched numbers on the cleaned splits, differences relative to EvoOptiGraph confound method quality with dataset cleaning and should not be read as a head-to-head comparison. The LLMOPT numbers are taken from the official repository (\url{https://github.com/antgroup/LLMOPT}) under the Single-Agent setting without self-correction, which matches our single-shot Pass@1 protocol.

\subsection{Evaluation Protocol and Remaining Caveats}
\label{app:eval-protocol}
Following common practice in natural-language-to-optimization modeling, we score Pass@1 by objective-value equality: a generated program is counted as correct when Gurobi returns an optimal objective that matches the benchmark reference within the relative-error tolerance $\tau_{\text{rel}}=10^{-4}$ used in training. This protocol is solver-verifiable and consistent with the RL correctness reward, but it does not directly inspect the algebraic formulation. Two programs with different constraint sets can still share the same optimal value, so a match does not certify structural identity with the intended model. Stronger structure-aware verification is left to future work.

ComplexOR and IndustryOR contain 18 and 42 instances, respectively. On these two smallest splits, a single additional correct or incorrect prediction changes Pass@1 by about $5.6$ and $2.4$ percentage points. Scores on these benchmarks therefore remain sensitive to a few instances and are most informative when read together with the larger splits and the macro-average.

\subsection{Mean and Standard Deviation}
\label{app:mean-std}
Table~\ref{tab:main} reports point estimates of Pass@1. For EvoOptiGraph, we additionally measure variability by training with 2 random seeds and evaluating each trained model on 3 independent runs. Table~\ref{tab:var} reports the mean $\pm$ standard deviation. Each entry is the mean over 3 evaluation runs and 2 training seeds. Avg.\ is the macro-average of the six per-benchmark means. The point estimates in Table~\ref{tab:main} fall within one standard deviation of these means on every benchmark. Standard deviations are largest on MAMO-ComplexLP and IndustryOR; ComplexOR is unchanged across the six runs. Baseline entries in Table~\ref{tab:main} remain the published point estimates.

\begin{table}[t]
\caption{Pass@1 of \texttt{EvoOptiGraph} (\%). Each entry is the mean $\pm$ standard deviation over 3 evaluation runs and 2 training seeds. Avg.\ is the macro-average of the six per-benchmark means.}
\label{tab:var}
\centering
\resizebox{\textwidth}{!}{%
\begin{tabular}{lcccccccc}
\toprule
\multirow{2}{*}{Model} & \multirow{2}{*}{Params} & \multirow{2}{*}{NL4OPT} & \multicolumn{2}{c}{MAMO} & \multirow{2}{*}{NLP4LP} & \multirow{2}{*}{CompOR} & \multirow{2}{*}{IndOR} & \multirow{2}{*}{Avg.} \\
\cmidrule(lr){4-5}
 & & & EasyLP & ComplexLP & & & & \\
\midrule
\texttt{EvoOptiGraph} & 8B & $96.9{\pm}0.4$ & $97.8{\pm}0.3$ & $67.7{\pm}2.0$ & $96.6{\pm}0.7$ & $50.0{\pm}0.0$ & $55.2{\pm}2.3$ & 77.4 \\
\bottomrule
\end{tabular}%
}
\end{table}

\section{Detailed Problem Descriptions for Illustrative Examples}
\label{app:example_details}

This appendix provides complete descriptions of the instances used in the crossover and mutation illustration (Figure~\ref{fig:crossover_mutation}) as well as additional examples generated by different crossover operations.

\paragraph{Problem A: Production Scheduling}
\textit{Natural language description.} A factory has 50 machine hours available. Two products can be manufactured:
\begin{itemize}
    \item Product 1 requires 2 hours per unit and yields a profit of 5 per unit.
    \item Product 2 requires 1 hour per unit and yields a profit of 3 per unit.
\end{itemize}
Both products can be produced in any non‑negative integer quantity. The objective is to maximize total profit.

\textit{Mathematical formulation.}
\[
\begin{aligned}
\max \quad & 5x_1 + 3x_2 \\
\text{s.t.} \quad & 2x_1 + x_2 \le 50, \\
& x_1, x_2 \ge 0,\ \text{integer}.
\end{aligned}
\]

\paragraph{Problem B: Knapsack}

\textit{Natural language description.} A knapsack has a capacity of 100. Two items are available:
\begin{itemize}
    \item Item 1 weighs 20 and has value 40.
    \item Item 2 weighs 25 and has value 60.
\end{itemize}
Each item can be selected at most once (0‑1 decision). The objective is to maximize total value.

\textit{Mathematical formulation.}
\[
\begin{aligned}
\max \quad & 40y_1 + 60y_2 \\
\text{s.t.} \quad & 20y_1 + 25y_2 \le 100, \\
& y_1, y_2 \in \{0,1\}.
\end{aligned}
\]

\paragraph{Problem C: Hybrid after Crossover}

\textit{Natural language description.} This instance is generated by combining the constraint subgraphs of Problem A and Problem B. It inherits both constraints: the machine‑hour limit from A and the knapsack capacity from B. The variables and objective coefficients are the union of those from A and B. In concrete terms, the factory has 50 machine hours and 100 units of raw material. Two integer products are available:
\begin{itemize}
    \item Product 1 requires 2 hours and yields a profit of 5.
    \item Product 2 requires 1 hour and yields a profit of 3.
\end{itemize}
Two binary tasks are available:
\begin{itemize}
    \item Task 1 consumes 20 units of material and gives a reward of 40.
    \item Task 2 consumes 25 units of material and gives a reward of 60.
\end{itemize}
Total profit is the sum of product profits and task rewards. The goal is to maximize total profit.

\textit{Mathematical formulation.}
\[
\begin{aligned}
\max \quad & 5x_1 + 3x_2 + 40y_1 + 60y_2 \\
\text{s.t.} \quad & 2x_1 + x_2 \le 50, \\
& 20y_1 + 25y_2 \le 100, \\
& x_1, x_2 \ge 0,\ \text{integer},\\
& y_1, y_2 \in \{0,1\}.
\end{aligned}
\]

\paragraph{Problem C': Mutation of Problem C}

\textit{Natural language description.} Problem C' is derived from Problem C by a parameter mutation: the raw material capacity is reduced from 100 to 90. All other aspects (variables, constraints, objective coefficients) remain unchanged.

\textit{Mathematical formulation.}
\[
\begin{aligned}
\max \quad & 5x_1 + 3x_2 + 40y_1 + 60y_2 \\
\text{s.t.} \quad & 2x_1 + x_2 \le 50, \\
& 20y_1 + 25y_2 \le 90, \\
& x_1, x_2 \ge 0,\ \text{integer},\\
& y_1, y_2 \in \{0,1\}.
\end{aligned}
\]

\noindent\textbf{More diverse instances via different crossover operations.}
Beyond the basic crossover shown in Problem~C (combining entire constraint subgraphs), our framework supports other crossover variants that produce even more diverse structures. We illustrate two such variants below: Problem~D demonstrates a simple exchange of constraint subgraphs between parent classes, while Problem~E shows a more sophisticated crossover that creates a novel mixed constraint.

\paragraph{Problem D: Another Crossover Example}

\textit{Natural language description.} This instance is generated by a different crossover operation: we exchange the entire constraint subgraph of Problem A with that of Problem B. The resulting offspring retains the variables and objective coefficients from A but adopts the constraint from B. In practical terms, the factory now has 100 units of raw material. Two integer products are available:
\begin{itemize}
    \item Product 1 consumes 20 units of raw material and yields a profit of 5 per unit.
    \item Product 2 consumes 25 units of raw material and yields a profit of 3 per unit.
\end{itemize}
Both products can be produced in any non‑negative integer quantity. The objective is to maximize total profit.

\textit{Mathematical formulation.}
\[
\begin{aligned}
\max \quad & 5x_1 + 3x_2 \\
\text{s.t.} \quad & 20x_1 + 25x_2 \le 100, \\
& x_1, x_2 \ge 0,\ \text{integer}.
\end{aligned}
\]

\textit{Discussion.} This operation produces a structurally distinct instance: a production scheduling problem transformed into a knapsack with integer items. It demonstrates the flexibility of our crossover to exchange subgraphs between different problem classes.

\paragraph{Problem E: Complex Crossover with a Novel Mixed Constraint}

\textit{Parent descriptions (same as A and B).} We use the same parents A and B as above.

\textit{Crossover operation.} We perform a three‑step crossover:
\begin{enumerate}
    \item Inherit the machine‑hour constraint from Parent A.
    \item Inherit the knapsack capacity constraint from Parent B.
    \item Create a new constraint by combining coefficients from both parents: take the coefficient of \(x_1\) from Parent A (2) and the coefficient of \(y_1\) from Parent B (20), and form the inequality \(2x_1 + 20y_1 \le 30\), where the right‑hand side is randomly generated (30). This new constraint couples an integer variable and a binary variable.
\end{enumerate}
The objective is the sum of both parents' objectives.

\textit{Resulting Problem E: Multi‑Constraint Hybrid.} The offspring contains three constraints and a mix of integer and binary variables.

\textit{Natural language description.} A factory has three independent resource limits:
\begin{itemize}
    \item \textbf{Machine hours:} 50 hours available. Two integer products can be produced: Product 1 requires 2 hours and yields a profit of 5 per unit; Product 2 requires 1 hour and yields a profit of 3 per unit.
    \item \textbf{Raw material:} 100 units available. Two binary tasks can be selected: Task 1 consumes 20 units of material and gives a reward of 40; Task 2 consumes 25 units of material and gives a reward of 60.
    \item \textbf{Catalyst constraint (mixed coupling):} Producing Product 1 consumes a rare catalyst (2 units per product); activating Task 1 (e.g., starting an auxiliary reactor) consumes 20 units of the same catalyst. The daily supply of catalyst is limited to 30 units. Hence, the production of Product 1 and the activation of Task 1 must satisfy \(2x_1 + 20y_1 \le 30\).
\end{itemize}
The goal is to maximize total profit, which is the sum of product profits and task rewards.

\textit{Mathematical formulation.}
\[
\begin{aligned}
\max \quad & 5x_1 + 3x_2 + 40y_1 + 60y_2 \\
\text{s.t.} \quad & 2x_1 + x_2 \le 50, \\
& 20y_1 + 25y_2 \le 100, \\
& 2x_1 + 20y_1 \le 30, \\
& x_1, x_2 \ge 0,\ \text{integer},\\
& y_1, y_2 \in \{0,1\}.
\end{aligned}
\]

\textit{Discussion.} This example demonstrates a highly flexible crossover that not only inherits complete constraints from parents but also creates entirely new constraints by mixing coefficients and variable types. The resulting instance has a richer structure (three constraints, mixed variable domains, a coupling between an integer and a binary variable) that would be impossible to obtain by simple parameter perturbation or by merely concatenating parent constraints. Such complex crossovers are crucial for generating structurally diverse training data, enabling the model to learn intricate trade‑offs between different resource types and decision variables. Our framework supports this kind of compositional crossover by allowing arbitrary compatible subgraphs (including partial coefficient vectors) to be exchanged or combined.

\begin{figure}[t]
    \centering
    \includegraphics[width=1\linewidth]{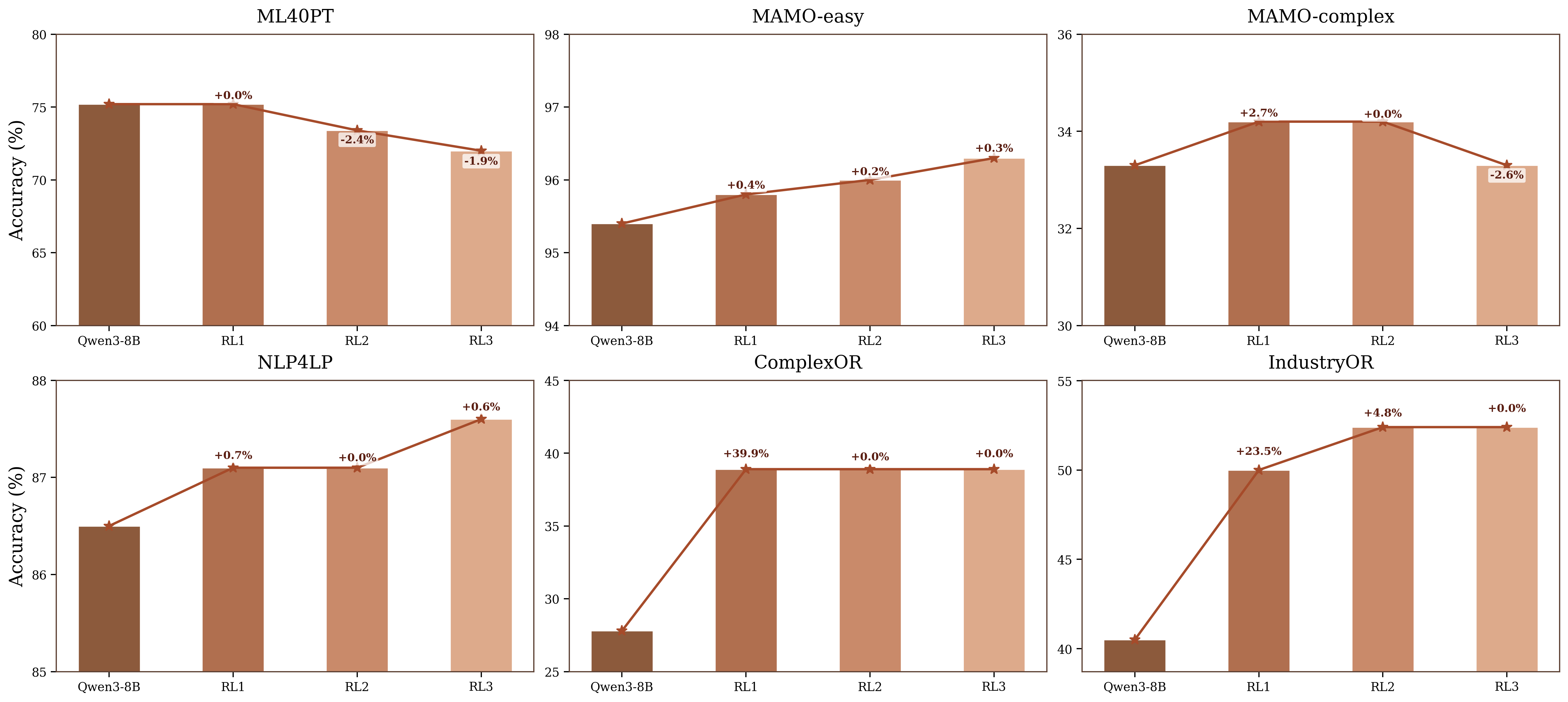}
    \caption{Accuracy trajectory of RL-only. The relative improvement of current iteration over the previous one is demonstrated on the corresponding bar.}
    \label{fig:RL}
\end{figure}

\begin{figure}[t]
    \centering
    \includegraphics[width=1\linewidth]{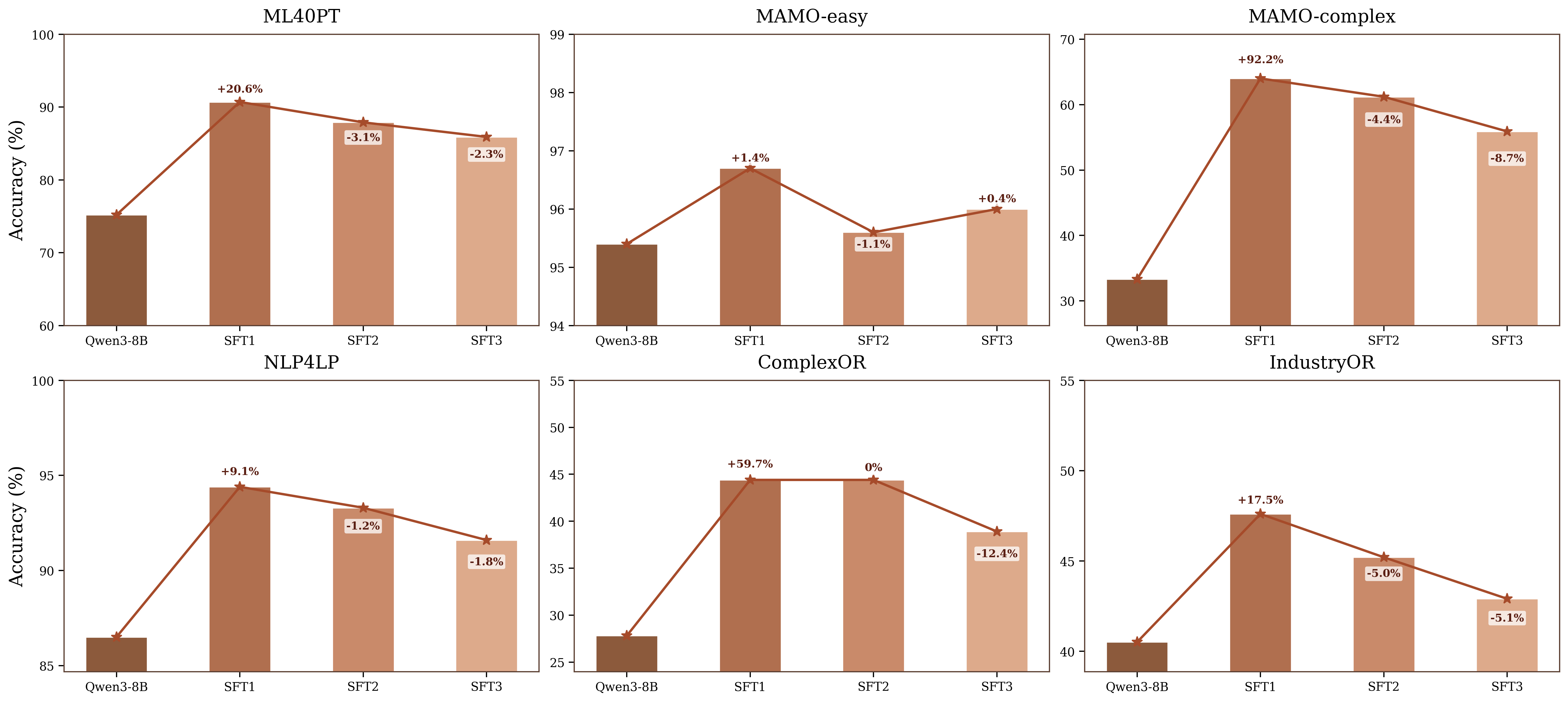}
    \caption{Accuracy trajectory of SFT-only. The relative improvement of current iteration over the previous one is demonstrated on the corresponding bar.}
    \label{fig:STF}
\end{figure}

\section{Details of Implementation}
\subsection{SFT \& RL}
For supervised fine-tuning (SFT), we use the \textsc{LlamaFactory} framework \cite{zheng2024llamafactoryunifiedefficientfinetuning} with a learning rate of \(1\times 10^{-4}\) for 3--5 epochs. We employ LoRA with rank \(32\), alpha \(64\), and dropout \(0.05\). The warm-up ratio is set to \(0.1\), and the optimizer is \texttt{adamw\_8bit}. Although some hyperparameters vary slightly across experiments, the overall setup remains similar. For reinforcement learning (RL), we use the \textsc{GRPOTrainer} implementation from the \textsc{TRL} library \cite{vonwerra2020trl}. Instead of full-parameter updates, we optimize LoRA adapters with the same LoRA hyperparameters as in SFT. We set the number of candidate generations per prompt to \(4\), the sampling temperature to \(0.7\), the number of training epochs to \(3\), and the learning rate to \(5\times 10^{-6}\).

\subsection{Back-translation Audit and Label Noise}
\label{app:audit}
We run periodic value-matching audits on sampled instances before they enter $\mathcal{D}_{\text{RL}}$. For each sampled instance, the generated natural-language problem is re-parsed and re-solved, and the instance is discarded when the recovered optimal value deviates from the graph's $f_{\text{opt}}$ by more than the proximity tolerance. On a randomly sampled audit set of 500 instances, 6.2\% failed this automatic check and were filtered out. To estimate the residual mismatch that remains after filtering, we manually inspected 100 post-filter instances and identified 2 genuine natural-language/ground-truth semantic mismatches (2\%). As with the benchmark protocol in Appendix~\ref{app:eval-protocol}, this automatic filter checks recovered objective values rather than formulation structure, so residual semantic errors of this kind can remain after filtering.

The staged reward is deliberately designed to limit exposure to label noise. Of the four reward terms, $R_{\text{process}}$ (execution) and $R_{\text{solvable}}$ (feasibility) are computed by running Gurobi on the generated code and are fully label-independent; only $R_{\text{proximity}}$ and $R_{\text{correctness}}$ depend on $f_{\text{opt}}$. Label noise can therefore corrupt only a bounded portion of the total reward.

\subsection{Genetic Algorithm}
\label{app:ga}
The weights in fitness function are set to $\beta = 0.3$, $\gamma = 0.3$, and $\delta$ follows a linear warm-up schedule from $0$ to $0.7$ over the first 30\% of generations (i.e., $\delta$ increases linearly during the warm-up phase and remains at $0.7$ thereafter). The GA uses a population size of $\max(30,\; 2N_{\text{target}})$, runs for 10 generations with tournament selection ($k=3$), crossover rate $0.8$, mutation rate $0.5$, per-coefficient mutation probability $0.10$ with Gaussian noise scale $0.12$, size mutation probability $0.20$, and elite ratio $15\%$. Feasibility is verified by Gurobi with a time limit of $300$\,s and MIP gap $0.01$.

\textbf{Cross-group crossover.}
Group pairs are selected for crossover only when their compatibility score
$s = 0.40\,s_{\text{vtype}} + 0.20\,s_{\text{density}} + 0.25\,s_{\text{size}} + 0.15\,s_{\text{sense}} \geq 0.60$,
where the four terms measure cosine similarity of variable-type distributions, constraint-matrix density difference, relative size difference, and optimization-direction consistency, respectively.
The operator exchanges contiguous blocks of constraint rows at a random split point $r \in [\lceil n/4 \rceil,\, \lfloor 3n/4 \rfloor]$: rows before $r$ come from one parent and rows after from the other, each carrying its full coefficient vector and bounds.
Variable columns are independently inherited per column from a randomly chosen parent, ensuring no orphaned variables.
If the offspring is infeasible, a repair step preserves the cross-group constraint matrix $A$ and resets $b_l, b_u$ around a randomly sampled feasible point $\mathbf{x}^*$ to restore feasibility.

\subsection{Instance Feasibility and Validity}
\label{app:validity}
GenBench-MILP evaluates generated MILP instances by feasibility as well as usefulness \cite{luo2026generatedinstancestrulyuseful}. Usefulness, measured by structural similarity to the 53 seed problem types, is reported in Section~\ref{sec:diversity}. We report structural validity and solver feasibility here.

Every offspring that enters the candidate pool is structurally valid. The genetic operators preserve validity: crossover keeps exactly one objective node and retains complete variable bounds and metadata, and a deterministic rule-based check and repair step runs before compilation. Every structurally valid candidate graph compiles into executable Gurobi Python code without syntax or solver API errors.

On a random sample of 1,000 compiled models, roughly 86.2\% are solved to feasibility or optimality within the $300$\,s time limit, 10.3\% are infeasible with a non-empty irreducible infeasible subset (IIS), and 3.5\% encounter solver timeouts. (Approximately 3\% of the infeasible samples are retained only for SFT and do not enter weakness analysis or RL.)

\subsection{XGBoost}
\label{app:xgboost}
An XGBoost classifier was trained to predict whether the solution outcome for each generated optimization instance was correct. The input representation comprised graph-derived descriptors computed from the optimization model structure, including statistics of the constraint matrix, objective vector, right-hand-side bounds, variable bounds, variable types, and instance-level metadata. In total, 102 features were constructed, as summarized in Table \ref{feature}. To improve robustness and reduce redundancy, feature selection was performed in three steps. First, features with zero variance were removed. Next, among feature pairs with absolute correlation greater than 0.95, the feature with lower mutual information with the target label was pruned. Finally, the top 20 features ranked by mutual information were retained. Retained features are then standardized by $z$-score (zero mean, unit variance) using statistics computed on that round's training split. The final XGBoost configuration used 500 boosting rounds, a maximum tree depth of 4, and a learning rate of 0.05. The subsample ratio and column sampling ratio per tree were set to 0.8 and 0.6, respectively. In addition, the minimum child weight was set to 10, $\gamma$ was set to 1.0, the L1 regularization coefficient was 1.0, and the L2 regularization coefficient was 5.0. The classifier was trained with the \texttt{binary:logistic} objective and evaluated primarily using the AUC metric.

The failure probe is retrained from scratch at every coevolution round on the current model's validation outcomes. Held-out AUC is typically around $0.88$ (e.g., $0.8836$ for the run in Figure~\ref{fig:example}). The resulting weakness vector remains a heuristic, correlational prioritization signal rather than a causal account of errors, and we do not report a detailed round-by-round stability analysis of $\mathbf{w}_t$.

\input{table/table3}

\definecolor{promptgray}{HTML}{A0A0A0}
\definecolor{promptbg}{HTML}{F0F0F0}
\definecolor{promptwhite}{HTML}{FFFFFF}
\definecolor{promptcyan}{HTML}{006D77}

\newtcolorbox{promptbox}[1]{
  enhanced,
  breakable,
  colback=promptwhite,
  colframe=promptgray,
  colbacktitle=promptbg,
  coltitle=black,
  coltext=promptcyan,
  title={#1},
  fonttitle=\bfseries,
  boxrule=0.9pt,
  arc=3mm,
  outer arc=3mm,
  left=12pt,
  right=12pt,
  top=10pt,
  bottom=10pt,
  toptitle=3pt,
  bottomtitle=3pt,
  lefttitle=10pt,
  righttitle=10pt,
}
\subsection{Prompt Templates}
\subsubsection{Baseline Prompt}
\begin{promptbox}{Baseline Prompt Template for optimization modeling}
\begin{Verbatim}[
  fontsize=\small,
  breaklines=true,
  breakanywhere=true,
  breaksymbolleft={},
  breaksymbolright={}
]
Below is an operations research question. Build a mathematical model and corresponding python code using `gurobipy` that appropriately addresses the question.

# Question:
{question}

# Response:
\end{Verbatim}
\end{promptbox}

\subsubsection{Reverse Data Generate Prompt}
\begin{promptbox}{Prompt for Cross-Group Example Generation}
\begin{Verbatim}[
  fontsize=\small,
  breaklines=true,
  breakanywhere=true,
  breaksymbolleft={},
  breaksymbolright={}
]
As an Operations Research Expert, your task is to create a SINGLE coherent optimization problem description that naturally combines two problem types into one unified scenario, with a CLEAR emphasis on one type as the primary framework.

## Primary Problem Type (DOMINANT): <DOM_SUBCLASS>

Mathematical Model:
<DOM_FORMULA>

Example:
<DOM_EXAMPLE>

## Secondary Problem Type (supplementary): <MIN_SUBCLASS>

Mathematical Model:
<MIN_FORMULA>

Example:
<MIN_EXAMPLE>

## Your Task:
Create ONE natural language description of an optimization problem that:
1. Presents a single, realistic scenario (e.g., a company, a logistics operation, a planning task)
2. Uses <DOM_SUBCLASS> as the PRIMARY framework and core storyline
3. Naturally weaves in <MIN_SUBCLASS> elements as SECONDARY considerations
4. The <DOM_SUBCLASS> aspects should be clearly dominant (roughly 60-70% of the problem), while <MIN_SUBCLASS> provides the remaining enrichment
5. Includes specific numerical parameters (costs, capacities, demands, etc.) to make it concrete
6. Clearly states what needs to be optimized
7. Describes all constraints naturally within the narrative
8. Uses appropriate domain terminology from both fields
9. Does NOT contain any mathematical formulas or variable names

For example, if the dominant type is "Knapsack" and secondary is "Bin Packing", you might describe a purchasing manager who primarily needs to select which products to buy within a budget, and additionally must arrange selected products into shipping containers.

## Required Output:
Provide ONLY the natural language problem description. No headers, no meta-commentary, no formulas. Just the problem scenario.
\end{Verbatim}
\end{promptbox}

\begin{promptbox}{Prompt for Single-Group Reverse Generation}
\begin{Verbatim}[
  fontsize=\small,
  breaklines=true,
  breakanywhere=true,
  breaksymbolleft={},
  breaksymbolright={}
]
As an Operations Research Expert, analyze the given LP data and generate a natural language description.

Input LP Data:
<LP_DATA>

Mathematical Model Reference:
<FORMULA>

Application Domain: <SUBCLASS>

Reference Example:
<EXAMPLE>

## Required Output:
Generate a clear, detailed natural language description of this optimization problem that:
- Describes a concrete, realistic scenario in the <SUBCLASS> domain
- States all decision variables and what they represent
- Specifies the objective (minimize/maximize) and what it means in the scenario
- Incorporates ALL constraints naturally within the narrative
- Includes ALL numerical parameters (coefficients, bounds, right-hand sides) exactly as they appear in the LP data
- Uses appropriate domain terminology
- Does NOT include any mathematical formulas, variable names like x0/x1, or LP file content

CRITICAL: Every numerical value in the LP data must appear somewhere in your description. Do not omit or approximate any parameter.

Provide ONLY the natural language description, nothing else.
\end{Verbatim}
\end{promptbox}

\begin{promptbox}{Prompt for Dominant Cross-Group Generation}
\begin{Verbatim}[
  fontsize=\small,
  breaklines=true,
  breakanywhere=true,
  breaksymbolleft={},
  breaksymbolright={}
]
As an Operations Research Expert, analyze the given LP data and generate a natural language description.

This optimization problem primarily belongs to "<DOM_SUBCLASS>" (approx. <DOM_RATIO>\%), with additional constraints from "<MIN_SUBCLASS>" (approx. <MIN_RATIO>\%).

Input LP Data:
<LP_DATA>

Primary Mathematical Model Reference (<DOM_SUBCLASS>):
<DOM_FORMULA>

Cross-Domain Reference Example (showing how <DOM_SUBCLASS> and <MIN_SUBCLASS> can be combined in one scenario):
<CROSS_EXAMPLE>

## Required Output:
Following the style of the cross-domain reference example above, generate a natural language description that:
- Frames the problem primarily as a <DOM_SUBCLASS> problem
- Naturally incorporates the <MIN_SUBCLASS>-style constraints as part of the same scenario
- Describes a concrete, realistic scenario
- States all decision variables and what they represent
- Specifies the objective (minimize/maximize) clearly
- Incorporates ALL constraints naturally within the narrative
- Includes ALL numerical parameters (coefficients, bounds, right-hand sides) EXACTLY as they appear in the LP data
- Does NOT include any mathematical formulas, variable names like x0/x1, or LP file content

CRITICAL: Every numerical value in the LP data must appear somewhere in your description. Do not omit or approximate any parameter. The description must be fully self-contained - a reader should be able to reconstruct the LP formulation from your description alone.

Provide ONLY the natural language description, nothing else.
\end{Verbatim}
\end{promptbox}

\begin{promptbox}{Prompt for Mixed Cross-Group Generation}
\begin{Verbatim}[
  fontsize=\small,
  breaklines=true,
  breakanywhere=true,
  breaksymbolleft={},
  breaksymbolright={}
]
As an Operations Research Expert, analyze the given LP data and generate a natural language description.

This optimization problem combines two domains:
- Primary: "<DOM_SUBCLASS>" (approx. <DOM_RATIO>\%)
- Secondary: "<MIN_SUBCLASS>" (approx. <MIN_RATIO>\%)

Input LP Data:
<LP_DATA>

Primary Mathematical Model Reference (<DOM_SUBCLASS>):
<DOM_FORMULA>

Cross-Domain Reference Example (showing how to combine <DOM_SUBCLASS> and <MIN_SUBCLASS> naturally):
<CROSS_EXAMPLE>

## Required Output:
Using the cross-domain reference example as a style guide, generate a natural language description that:
- Creates a unified, realistic scenario combining aspects of both <DOM_SUBCLASS> and <MIN_SUBCLASS>
- The primary framing should be from the <DOM_SUBCLASS> perspective
- <MIN_SUBCLASS> constraints should be woven in naturally
- States all decision variables and what they represent
- Specifies the objective clearly
- Incorporates ALL constraints within the narrative
- Includes ALL numerical parameters EXACTLY as in the LP data
- Does NOT include any mathematical formulas, variable names like x0/x1, or LP file content

CRITICAL: Every numerical value in the LP data must appear somewhere in your description. Do not omit or approximate any parameter. The description must be fully self-contained - a reader should be able to reconstruct the LP formulation from your description alone.

Provide ONLY the natural language description, nothing else.
\end{Verbatim}
\end{promptbox}

\begin{promptbox}{Prompt for Fusion Cross-Group Generation}
\begin{Verbatim}[
  fontsize=\small,
  breaklines=true,
  breakanywhere=true,
  breaksymbolleft={},
  breaksymbolright={}
]
As an Operations Research Expert, analyze the given LP data and generate a natural language description.

This optimization problem equally combines:
- "<DOM_SUBCLASS>" (approx. <DOM_RATIO>\%)
- "<MIN_SUBCLASS>" (approx. <MIN_RATIO>\%)

Input LP Data:
<LP_DATA>

Cross-Domain Reference Example (demonstrating how <DOM_SUBCLASS> and <MIN_SUBCLASS> merge into one scenario):
<CROSS_EXAMPLE>

## Required Output:
Using the cross-domain reference example as a style guide, generate a natural language description that:
- Creates a single coherent, realistic scenario integrating BOTH <DOM_SUBCLASS> and <MIN_SUBCLASS> equally
- Both problem aspects should be equally prominent in the narrative
- States all decision variables and what they represent
- Specifies the objective clearly
- Incorporates ALL constraints within the narrative
- Includes ALL numerical parameters EXACTLY as in the LP data
- Does NOT include any mathematical formulas, variable names like x0/x1, or LP file content

CRITICAL: Every numerical value in the LP data must appear somewhere in your description. Do not omit or approximate any parameter. The description must be fully self-contained - a reader should be able to reconstruct the LP formulation from your description alone.

Provide ONLY the natural language description, nothing else.
\end{Verbatim}
\end{promptbox}

\begin{promptbox}{Prompt for Self-Critique}
\begin{Verbatim}[
  fontsize=\small,
  breaklines=true,
  breakanywhere=true,
  breaksymbolleft={},
  breaksymbolright={}
]
As an Operations Research Expert, rigorously verify if the generated problem description accurately and completely represents the LP data.

Input LP Data:
<LP_DATA>

Generated Problem Description:
<DESCRIPTION>

## Verification Checklist:
1. Count decision variables in LP data vs description - do they match?
2. For EACH variable: are bounds (lower, upper) correctly stated?
3. For EACH variable: is the type (integer, binary, continuous) correctly described?
4. Is the objective function direction (min/max) correct?
5. For EACH objective coefficient: does the description include the exact value?
6. Count constraints in LP data vs description - do they match?
7. For EACH constraint: are ALL coefficients exactly correct?
8. For EACH constraint: are the right-hand side values (bounds) exactly correct?
9. Are there any numerical values in the LP data that are missing from the description?
10. Are there any numerical values in the description that do NOT appear in the LP data?

## Required Output:
If ALL checks pass perfectly:
"Complete Instance"

If ANY discrepancy exists:
"Incomplete Instance:
[List EVERY specific discrepancy with exact values from LP data that are wrong or missing]"
\end{Verbatim}
\end{promptbox}

\begin{promptbox}{Prompt for Self-Refinement}
\begin{Verbatim}[
  fontsize=\small,
  breaklines=true,
  breakanywhere=true,
  breaksymbolleft={},
  breaksymbolright={}
]
As an Operations Research Expert, fix the problem description based on the 
criticism.

Criticism:
<CRITICISM>

If the criticism says "Complete Instance", output exactly: "Nothing need to refine"

Otherwise, fix ALL identified issues:

LP Data (ground truth):
<LP_DATA>

Scenario Context: <SCENARIO_CONTEXT>

Current Description (contains errors):
<INITIAL_DESCRIPTION>

## Required Output:
If criticism is "Complete Instance":
Output "Nothing need to refine"

Otherwise, generate a CORRECTED description that:
- Fixes every single discrepancy identified in the criticism
- Maintains the same narrative style and scenario
- Includes ALL numerical parameters EXACTLY as in the LP data
- Does NOT include any mathematical formulas or LP file content
- Is a complete, standalone description (not a patch or diff)

Output ONLY the corrected natural language description, nothing else.
\end{Verbatim}
\end{promptbox}

\section{Comparison with Text-Only Evolution}
\label{app:text-only}
We compare graph-based evolution with a text-only evolution baseline based on MILP-Evolve \citep{li2024towards}, an LLM-based evolutionary framework that generates diverse MILP classes and an unlimited number of instances. In this baseline, an LLM mutates the natural-language description or the solver code directly. EvoOptiGraph edits the bipartite constraint graph and uses the weakness vector to guide which instances are generated. The SFT warm-up, RL recipe, and per-round data budget are otherwise identical. Text-only evolution does not use the weakness vector.

Table~\ref{tab:text-only} reports Pass@1 under this matched setting. Text-only evolution is lower on every benchmark, and the macro-average gap is 8.4 percentage points (78.1 vs.\ 69.7). Two factors account for the difference. First, because text-only evolution has no weakness vector, new instances are not steered toward the model's current failure modes, so the training data are less targeted. Second, text-level mutations often produce semantically ambiguous constraints or invalid formulations that fail Gurobi compilation. The required re-sampling lowers the effective data yield and reduces the diversity of the instances that remain.

\begin{table}[t]
\caption{Graph-based evolution versus text-only evolution under the same SFT warm-up, RL recipe, and per-round data budget. Pass@1 accuracy (\%) on six OR benchmarks.}
\label{tab:text-only}
\centering
\small
\setlength{\tabcolsep}{3.5pt}
\renewcommand{\arraystretch}{0.9}
\begin{tabular}{lccccccc}
\toprule
\multirow{2}{*}{Setting} & \multirow{2}{*}{NL4OPT} & \multicolumn{2}{c}{MAMO} & \multirow{2}{*}{NLP4LP} & \multirow{2}{*}{CompOR} & \multirow{2}{*}{IndOR} & \multirow{2}{*}{Avg.} \\
\cmidrule(lr){3-4}
 & & EasyLP & ComplexLP & & & & \\
\midrule
Text-only (MILP-Evolve) & 88.3 & 96.1 & 50.5 & 91.0 & 44.4 & 47.6 & 69.7 \\
EvoOptiGraph & 96.7 & 98.0 & 69.4 & 97.2 & 50.0 & 57.1 & 78.1 \\
\bottomrule
\end{tabular}
\end{table}

\section{Supplementary Technical Details}
\subsection{Formal Definition of Graph Edit Distance (GED) and Its Application in This Paper}

Graph Edit Distance (GED) is a classic measure of structural dissimilarity between two graphs \citep{gao2010survey, gasse2019exact, fan2023smart}. For two graphs $G_1 = (V_1, E_1)$ and $G_2 = (V_2, E_2)$, GED is defined as the minimum cost of a sequence of edit operations (insertion, deletion, and substitution of nodes and edges) that transforms $G_1$ into $G_2$. Its mathematical definition is:
\[
\operatorname{GED}(G_1, G_2) = \min_{(o_1, \dots, o_k) \in \mathcal{P}(G_1, G_2)} \sum_{i=1}^k c(o_i),
\]
where $c(o_i)$ is the cost of operation $o_i$, and $\mathcal{P}(G_1, G_2)$ is the set of all valid sequences of operations.

In this paper, we adopt unit costs (all operations have cost 1) and compute GED efficiently using a bipartite matching approximation algorithm \citep{blumenthal2020comparing}. In Section~\ref{sec:data-evolution}, GED is used to compute the diversity term $\tilde{f}_{\text{div}}(\mathcal{G})$, i.e., the normalized average GED between the current individual and other individuals in the population. A larger GED indicates a more significant structural difference, thereby encouraging the evolution to explore novel graph structures. This design ensures structural diversity of the population, providing a foundation for subsequent difficulty selection and weakness guidance.

\subsection{How the Weakness Signal Guides Generation}
\label{app:weakness-guidance}
The weakness signal guides \textsc{EvoGenerate} through the weakness-alignment term of the fitness function. In each coevolution round, the current model is evaluated on the held-out validation set to obtain its error set \(\mathcal{E}_t\). An XGBoost classifier is trained to predict correctness from the structural feature vector \(\phi(\mathcal{G})\), and SHAP values on the error cases form the weakness vector
\[
w_k = \frac{1}{|\mathcal{E}_t|} \sum_{i \in \mathcal{E}_t} \mathrm{SHAP}_{i,k},
\]
where each coordinate is the direction along which a structural feature is most associated with failure.

This vector enters generation through the fitness
\[
f(\mathcal{G}) = \beta \tilde{f}_{\mathrm{diff}} + \gamma \tilde{f}_{\mathrm{div}} + \delta \tilde{f}_{\mathrm{weak}}, \qquad \tilde{f}_{\mathrm{weak}} = \cos\bigl(\phi(\mathcal{G}), \mathbf{w}_t\bigr).
\]
The weight \(\delta\) follows the schedule in Section~\ref{app:ga}: it increases linearly from \(0\) to \(0.7\) over the first \(30\%\) of generations and stays at \(0.7\) thereafter.

A candidate whose structure is closer to the model's current failure regime receives a higher fitness, so tournament selection is more likely to keep and propagate it. The resulting dataset \(\mathcal{D}_{\mathrm{RL}}\) is therefore biased toward the model's structural blind spots. These instances drive the GRPO update. The next round re-diagnoses the updated model and produces \(\mathbf{w}_{t+1}\), so the training distribution co-evolves with the model.

This construction remains heuristic. Cosine alignment with a SHAP-aggregated vector is a convenient targeting mechanism rather than a guarantee that the selected instances cause the observed failures. The probe itself is specified in Appendix~\ref{app:xgboost}.

\subsection{t-SNE Dimensionality Reduction and Its Application in Weakness Analysis}

\paragraph{Brief Introduction to t-SNE}

t-SNE is a nonlinear dimensionality reduction technique that preserves local structure by minimizing the KL divergence between probability distributions in the high-dimensional and low-dimensional spaces \citep{maaten2008visualizing}. It defines high-dimensional similarities $p_{ij}$ and low-dimensional similarities $q_{ij}$ and optimizes the low-dimensional coordinates via gradient descent.

\paragraph{Application in Weakness Analysis}

We use t-SNE to visualize the feature vectors $\phi(\mathcal{G})$ to validate the effectiveness of the extracted features. The procedure is as follows:

\begin{enumerate}
    \item Select correctly and incorrectly solved instances from the validation set $V_{val}$.
    \item Extract the $d$-dimensional feature vector $\phi(\mathcal{G})$ for each instance (see Section~3.3 for details).
    \item Apply t-SNE to project the features into two dimensions for visualization.
\end{enumerate}

\begin{figure}[t]
    \centering
    \includegraphics[width=1\linewidth]{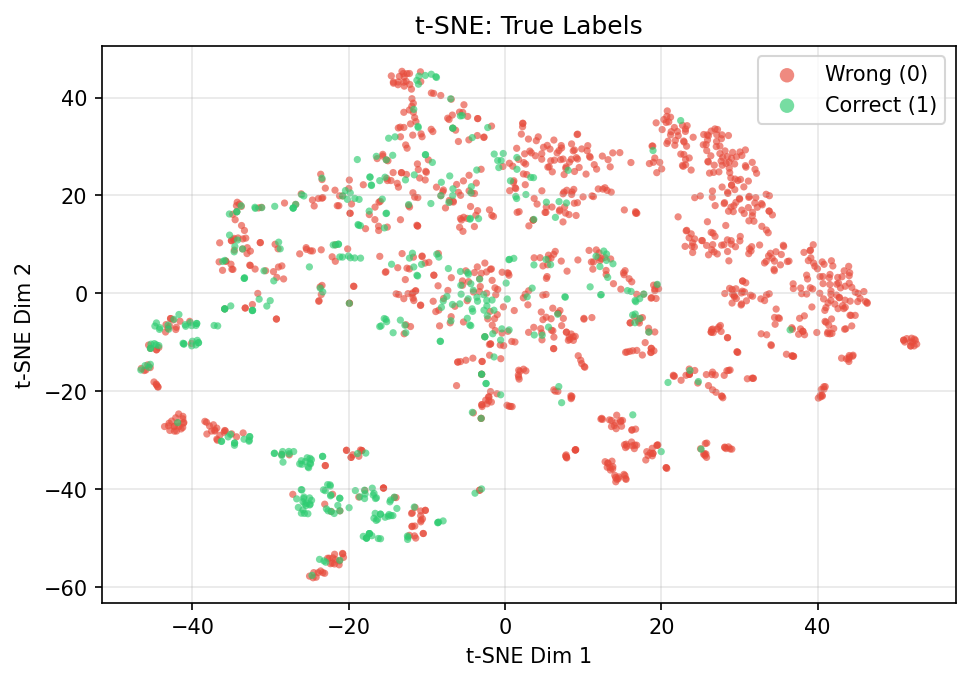}
    \caption{An Example of t-SNE Feature Visualization}
    \label{fig:example}
\end{figure}

In the resulting plot, correctly and incorrectly solved instances are marked with different colors. Figure \ref{fig:example} shows one representative example from a particular run. If the two classes exhibit visible clustering patterns, this provides visual evidence that the feature vector $\phi(\mathcal{G})$ captures structural information relevant to model errors. For the example shown in Figure \ref{fig:example}, the corresponding classifier achieves an AUC of 0.8836.

%% file: table/table3.tex


\definecolor{groupbg}{gray}{0.92}

\setlength{\LTleft}{0pt}
\setlength{\LTright}{0pt}

\begin{longtable}{%
  >{\ttfamily\small}p{0.30\textwidth}
  p{0.52\textwidth}
  >{\centering\arraybackslash\small}p{0.12\textwidth}
}
\label{feature} \\ 
\toprule
\normalfont\textbf{Feature Name} &
\textbf{Description} &
\textbf{Type} \\
\midrule
\endfirsthead

\toprule
\normalfont\textbf{Feature Name} &
\textbf{Description} &
\textbf{Type} \\
\midrule
\endhead

\midrule
\multicolumn{3}{r}{\small\textit{Continued on next page\ldots}} \\
\endfoot

\bottomrule
\endlastfoot

\multicolumn{3}{l}{%
  \cellcolor{groupbg}\textbf{A\quad Problem Dimensions}} \\[1pt]

n\_vars
  & Number of decision variables ($n_{\text{vars}}$)
  & Count \\

n\_cons
  & Number of constraints ($n_{\text{cons}}$)
  & Count \\

var\_con\_ratio
  & Ratio of variables to constraints
    ($n_{\text{vars}} / n_{\text{cons}}$)
  & Ratio \\

log\_size
  & Logarithm of total problem size:
    $\log(1 + n_{\text{vars}} \times n_{\text{cons}})$
  & Cont. \\[4pt]

\multicolumn{3}{l}{%
  \cellcolor{groupbg}\textbf{B\quad Variable Type Composition}} \\[1pt]

prop\_continuous
  & Proportion of continuous (\texttt{C}) variables
  & Ratio \\

prop\_binary
  & Proportion of binary (\texttt{B}) variables
  & Ratio \\

prop\_integer
  & Proportion of general integer (\texttt{I}) variables
  & Ratio \\

is\_pure\_lp
  & Binary indicator: problem is a pure LP
    (no integer or binary variables)
  & Binary \\

has\_integer
  & Binary indicator: problem contains at least one
    integer or binary variable
  & Binary \\[4pt]

\multicolumn{3}{l}{%
  \cellcolor{groupbg}\textbf{C\quad Constraint Type}} \\[1pt]

prop\_equality
  & Proportion of equality constraints
    ($b_l \approx b_u$)
  & Ratio \\[4pt]

\multicolumn{3}{l}{%
  \cellcolor{groupbg}\textbf{D\quad Sparsity and Non-zero Structure}} \\[1pt]

sparsity
  & Sparsity of constraint matrix $\mathbf{A}$:
    proportion of zero entries
  & Ratio \\

avg\_nnz\_per\_row
  & Mean number of non-zero entries per constraint row
  & Cont. \\

std\_nnz\_row
  & Standard deviation of non-zero counts per row
  & Cont. \\

std\_nnz\_col
  & Standard deviation of non-zero counts per column
  & Cont. \\[4pt]

\multicolumn{3}{l}{%
  \cellcolor{groupbg}\textbf{E\quad Coefficient Statistics (Matrix $\mathbf{A}$)}} \\[1pt]

log\_max\_coeff
  & $\log(1 + \max|\mathbf{A}_{ij}|)$ for non-zero entries
  & Cont. \\

coeff\_cv
  & Coefficient of variation of non-zero coefficients in $\mathbf{A}$:
    $\sigma / (\mu + \epsilon)$
  & Cont. \\

prop\_unit\_coeff
  & Proportion of non-zero entries equal to $\pm 1$
  & Ratio \\

has\_frac\_coeff
  & Binary indicator: at least one fractional (non-integer)
    coefficient exists in $\mathbf{A}$
  & Binary \\

coeff\_range\_log
  & Log-scale coefficient range in $\mathbf{A}$:
  
    $\log\!\left(1 + \max|a_{ij}| / (\min|a_{ij}| + \epsilon)\right)$
  & Cont. \\[4pt]

\multicolumn{3}{l}{%
  \cellcolor{groupbg}\textbf{F\quad Objective Coefficient Statistics}} \\[1pt]

prop\_zero\_obj
  & Proportion of zero entries in objective vector $\mathbf{c}$
  & Ratio \\

log\_max\_obj
  & $\log(1 + \max|\mathbf{c}_j|)$ for non-zero objective coefficients
  & Cont. \\

obj\_cv
  & Coefficient of variation of non-zero objective coefficients
  & Cont. \\[4pt]

\multicolumn{3}{l}{%
  \cellcolor{groupbg}\textbf{G\quad Sign Structure}} \\[1pt]

prop\_neg\_coeff
  & Proportion of negative entries in $\mathbf{A}$
  & Ratio \\

prop\_mixed\_sign\_rows
  & Proportion of constraint rows with both positive
    and negative coefficients
  & Ratio \\[4pt]

\multicolumn{3}{l}{%
  \cellcolor{groupbg}\textbf{H\quad Variable Bounds}} \\[1pt]

prop\_finite\_ub
  & Proportion of variables with a finite upper bound
  & Ratio \\

avg\_bound\_width
  & Mean width of finite variable bound intervals:
    $\overline{(u_j - l_j)}$ for bounded variables
  & Cont. \\

prop\_fixed\_vars
  & Proportion of variables fixed at a single value
    ($l_j = u_j$)
  & Ratio \\[4pt]

\multicolumn{3}{l}{%
  \cellcolor{groupbg}\textbf{I\quad Right-Hand Side Statistics}} \\[1pt]

log\_max\_rhs
  & $\log(1 + \max|b|)$ over all finite RHS values
  & Cont. \\

rhs\_cv
  & Coefficient of variation of finite RHS values
  & Cont. \\[4pt]

\multicolumn{3}{l}{%
  \cellcolor{groupbg}\textbf{J\quad Graph-Theoretic Properties}} \\[1pt]

graph\_density
  & Density of the bipartite constraint--variable graph:
    $\text{nnz}/(n_{\text{cons}} \times n_{\text{vars}})$
  & Ratio \\

degree\_std
  & Standard deviation of node degrees in the bipartite graph
  & Cont. \\

degree\_skewness
  & Skewness of the node degree distribution
  & Cont. \\[4pt]

\multicolumn{3}{l}{%
  \cellcolor{groupbg}\textbf{K\quad Objective--Constraint Geometric Similarity}} \\[1pt]

obj\_cons\_max\_cos
  & Maximum absolute cosine similarity between
    $\mathbf{c}$ and any constraint row of $\mathbf{A}$
  & Cont. \\

obj\_cons\_mean\_cos
  & Mean cosine similarity between $\mathbf{c}$
    and constraint rows of $\mathbf{A}$
  & Cont. \\

obj\_cons\_std\_cos
  & Standard deviation of cosine similarities
    between $\mathbf{c}$ and constraint rows
  & Cont. \\[4pt]

\multicolumn{3}{l}{%
  \cellcolor{groupbg}\textbf{L\quad Linear-Algebraic Properties}} \\[1pt]

rank\_ratio
  & Numerical rank ratio of $\mathbf{A}$:
  
    $\text{rank}(\mathbf{A}) / \min(n_{\text{cons}}, n_{\text{vars}})$
  & Ratio \\

row\_norm\_cv
  & Coefficient of variation of Euclidean row norms of $\mathbf{A}$
  & Cont. \\

prop\_tight\_cons
  & Proportion of constraints with a narrow feasible activity
    range (tightness ratio $< 0.1$)
  & Ratio \\

mean\_coupling
  & Mean number of shared variables between all pairs
    of constraints (off-diagonal of $\mathbf{A}\mathbf{A}^\top$)
  & Cont. \\[4pt]

\multicolumn{3}{l}{%
  \cellcolor{groupbg}\textbf{M\quad Connectivity}} \\[1pt]

n\_components
  & Number of connected components in the
    bipartite constraint--variable graph
  & Count \\

is\_connected
  & Binary indicator: the bipartite graph is connected
    (single component)
  & Binary \\[4pt]

\multicolumn{3}{l}{%
  \cellcolor{groupbg}\textbf{N\quad NLP Replacement: Problem Scale Proxies}} \\[1pt]

graph\_size
  & Total number of entries in $\mathbf{A}$:
    $n_{\text{cons}} \times n_{\text{vars}}$
  & Count \\

graph\_nnz
  & Total number of non-zero entries in $\mathbf{A}$
  & Count \\

graph\_n\_active\_cons
  & Number of constraints with at least one non-zero coefficient
  & Count \\

graph\_median\_coeff
  & Median absolute value of non-zero entries in $\mathbf{A}$
  & Cont. \\

graph\_mean\_nnz\_per\_col
  & Mean number of non-zero entries per variable column
  & Cont. \\

graph\_unique\_coeff\_ratio
  & Ratio of distinct non-zero values to total non-zeros in $\mathbf{A}$
  & Ratio \\[4pt]

\multicolumn{3}{l}{%
  \cellcolor{groupbg}\textbf{O\quad NLP Replacement: RHS Numerical Properties}} \\[1pt]

graph\_finite\_rhs\_count
  & Total count of finite values among all $b_l$ and $b_u$ bounds
  & Count \\

graph\_max\_finite\_abs
  & Maximum absolute value across all finite model parameters
    ($\mathbf{A}$, $\mathbf{c}$, $b_l$, $b_u$, $lb$, $ub$)
  & Cont. \\

graph\_has\_frac\_rhs
  & Binary indicator: at least one fractional RHS value exists
  & Binary \\

graph\_has\_neg\_rhs
  & Binary indicator: at least one negative RHS value exists
  & Binary \\

graph\_distinct\_rhs
  & Number of distinct RHS values across all bounds
  & Count \\

graph\_rhs\_range
  & Range of RHS values: $\max(b) - \min(b)$ over all finite bounds
  & Cont. \\[4pt]

\multicolumn{3}{l}{%
  \cellcolor{groupbg}\textbf{P\quad NLP Replacement: Constraint and Variable Type Counts}} \\[1pt]

graph\_int\_var\_count
  & Number of general integer (\texttt{I}) variables
  & Count \\

graph\_bin\_var\_count
  & Number of binary (\texttt{B}) variables
  & Count \\

graph\_is\_min
  & Binary indicator: the objective is a minimization
  & Binary \\

graph\_is\_max
  & Binary indicator: the objective is a maximization
  & Binary \\

graph\_leq\_cons\_count
  & Number of $\leq$ constraints
    (finite $b_u$, infinite $b_l$)
  & Count \\

graph\_geq\_cons\_count
  & Number of $\geq$ constraints
    (finite $b_l$, infinite $b_u$)
  & Count \\

graph\_eq\_cons\_count
  & Number of equality constraints ($b_l \approx b_u$)
  & Count \\

graph\_zero\_row\_count
  & Number of all-zero rows in $\mathbf{A}$
    (structurally redundant constraints)
  & Count \\

graph\_mixed\_sign\_count
  & Number of constraints containing both positive
    and negative coefficients
  & Count \\

graph\_linking\_pair\_count
  & Number of constraints with exactly 2 non-zero entries
    (binary linking constraints)
  & Count \\

graph\_frac\_coeff\_count
  & Number of fractional (non-integer) coefficients in $\mathbf{A}$
  & Count \\

graph\_obj\_range\_ratio
  & Ratio of maximum to minimum absolute objective coefficient:
    $\max|c_j| / (\min|c_j| + \epsilon)$
  & Cont. \\

graph\_unit\_row\_count
  & Number of constraint rows whose non-zero entries
    are all $\pm 1$
  & Count \\

graph\_range\_cons\_count
  & Number of range constraints with both finite and
    distinct $b_l$ and $b_u$ ($|b_u - b_l| \geq \epsilon$)
  & Count \\

graph\_obj\_density
  & Proportion of non-zero entries in objective vector $\mathbf{c}$
  & Ratio \\

graph\_singleton\_count
  & Number of constraints with exactly 1 non-zero entry
    (single-variable constraints)
  & Count \\

graph\_dense\_row\_count
  & Number of fully-dense rows in $\mathbf{A}$
    (all $n_{\text{vars}}$ variables appear)
  & Count \\[4pt]

\multicolumn{3}{l}{%
  \cellcolor{groupbg}\textbf{Q\quad NLP Replacement: Constraint Structural Diversity}} \\[1pt]

graph\_cons\_diversity
  & Number of distinct non-zero-count patterns
    (cardinalities) among constraint rows
  & Count \\[4pt]

\multicolumn{3}{l}{%
  \cellcolor{groupbg}\textbf{R\quad NLP Replacement: Code-Level Structural Proxies}} \\[1pt]

graph\_total\_coeff\_sum
  & Sum of absolute values of all entries in $\mathbf{A}$:
    $\sum_{i,j}|a_{ij}|$
  & Cont. \\

graph\_n\_rows
  & Number of constraint rows in the model
  & Count \\

graph\_n\_cols
  & Number of variable columns in the model
  & Count \\

graph\_bounded\_cons\_count
  & Number of constraints bounded on at least one side
    (finite $b_l$ or $b_u$)
  & Count \\

graph\_has\_identical\_rows
  & Binary indicator: at least two rows of $\mathbf{A}$ share
    an identical sign pattern
  & Binary \\

graph\_max\_col\_nnz
  & Maximum number of non-zero entries in any
    single variable column
  & Count \\

graph\_has\_bigM
  & Binary indicator: Big-M pattern detected
    (a coefficient exceeds $50\times$ the median non-zero value)
  & Binary \\

graph\_vtype\_diversity
  & Number of distinct variable types present
    (\texttt{C}, \texttt{I}, \texttt{B})
  & Count \\

graph\_obj\_nnz\_count
  & Number of non-zero entries in objective vector $\mathbf{c}$
  & Count \\

graph\_all\_finite\_ub
  & Binary indicator: all variables have finite upper bounds
  & Binary \\

graph\_var\_group\_count
  & Number of distinct variable subsets
    (support sets) appearing across constraint rows
  & Count \\

graph\_block\_mixing\_ratio
  & Proportion of constraints whose non-zeros span
    both the left and right halves of the variable index set
  & Ratio \\

graph\_n\_var\_names
  & Number of named decision variables in the model
  & Count \\

graph\_n\_con\_names
  & Number of named constraints in the model
  & Count \\

graph\_avg\_row\_abs\_sum
  & Mean sum of absolute coefficients per constraint row:
    $\overline{\sum_j|a_{ij}|}$
  & Cont. \\

graph\_max\_var\_participation
  & Maximum number of constraints in which
    any single variable appears
  & Count \\

graph\_total\_distinct\_values
  & Total number of distinct numeric values
    across $\mathbf{A}$, $\mathbf{c}$, $b_l$, $b_u$, $lb$, $ub$
  & Count \\[4pt]

\multicolumn{3}{l}{%
  \cellcolor{groupbg}\textbf{S\quad NLP Replacement: Internal Consistency Ratios}} \\[1pt]

graph\_rhs\_to\_cons\_ratio
  & Ratio of finite RHS count to number of constraints
  & Ratio \\

graph\_rhs\_coeff\_mag\_ratio
  & Ratio of mean absolute RHS to mean absolute
    coefficient in $\mathbf{A}$
  & Cont. \\

graph\_nnz\_to\_cons\_ratio
  & Ratio of total non-zeros in $\mathbf{A}$
    to number of constraints
  & Cont. \\

graph\_active\_cons\_ratio
  & Proportion of constraints with at least one
    non-zero coefficient
  & Ratio \\

graph\_obj\_A\_value\_overlap
  & Fraction of distinct objective coefficient values
    that also appear as entries in $\mathbf{A}$
  & Ratio \\

graph\_A\_only\_value\_count
  & Number of distinct numeric values appearing
    in $\mathbf{A}$ but not in $\mathbf{c}$
  & Count \\

graph\_c\_only\_value\_count
  & Number of distinct numeric values appearing
    in $\mathbf{c}$ but not in $\mathbf{A}$
  & Count \\

graph\_col\_jaccard
  & Mean Jaccard similarity of constraint participation sets
    over sampled pairs of variable columns
  & Ratio \\

graph\_complexity\_score
  & Composite complexity score:
    $n_{\text{cons}} \times \text{diversity} + \text{finite RHS count}$
  & Cont. \\

graph\_structural\_score
  & Structural complexity score:
    bounded constraints $+ n_{\text{vars}} + 2\cdot\mathbf{1}[\text{mixed}]
    + \text{group count}$
  & Cont. \\

graph\_complexity\_diff
  & Difference between composite and structural complexity: $\texttt{graph\_complexity\_score} - \texttt{graph\_structural\_score}$
  & Cont. \\[4pt]

\multicolumn{3}{l}{%
  \cellcolor{groupbg}\textbf{T\quad Instance Metadata}} \\[1pt]

problem\_type\_enc
  & Label-encoded integer representing the
    problem category (e.g.\ LP, MIP, MIQP)
  & Int. \\

inst\_n\_vars
  & Number of variables parsed from the instance identifier
  & Count \\

inst\_n\_cons
  & Number of constraints parsed from the instance identifier
  & Count \\

\end{longtable}